\documentclass[runningheads]{llncs}
\usepackage{graphicx}

\usepackage{tikz}
\usepackage{comment}
\usepackage{amsmath,amssymb,bm,stmaryrd} 
\usepackage{color}
\usepackage{enumitem}
\usepackage{subcaption}
\usepackage{xcolor}
\usepackage{multirow}
\usepackage[normalem]{ulem}
\usepackage{sidecap}
\usepackage{url}
\usepackage{footnote}
\usepackage{booktabs}

\usepackage[para,online,flushleft]{threeparttable}
\newcommand{\etal}{\textit{et al}.}
\newcommand{\ie}{\textit{i}.\textit{e}.}
\newcommand{\eg}{\textit{e}.\textit{g}.}

\usepackage{pifont}
\newcommand{\resnet}{ResNet}
\newcommand{\deit}{DeiT}
\newcommand{\beit}{BEiT}
\newcommand{\vit}{ViT}
\newcommand{\swin}{Swin}
\newcommand{\convnext}{ConvNeXt}

\newcommand{\trans}{Transformers}
\newcommand{\xmark}{\ding{55}}%
\newcommand*{\affmark}[1][*]{\textsuperscript{#1}}
\newcommand{\rom}[1]{\uppercase\expandafter{\romannumeral #1\relax}}

\begin{document}
\pagestyle{headings}
\mainmatter

\def\ECCVSubNumber{2862}  

\title{A Broad Study of Pre-training for Domain Generalization and Adaptation} 


\author{Donghyun Kim\inst{2} \and Kaihong Wang\inst{1} \and Stan Sclaroff\inst{1} \and Kate Saenko\inst{1,2}}
\institute{\affmark[1]Dept. CS, Boston University, \affmark[2]MIT-IBM Watson AI Lab\\
	\email{\{{donhk, kaiwkh, sclaroff, saenko\}@bu.edu}}}
\titlerunning{A Broad Study of Pre-training for Domain Transfer}	
\authorrunning{D. Kim \etal}

\maketitle

\begin{abstract}
Deep models must learn robust and transferable representations in order to perform well on new domains. While domain transfer methods (\eg, domain adaptation, domain generalization) have been proposed to learn transferable representations across domains, they are typically applied to ResNet backbones pre-trained on ImageNet. Thus, existing works pay little attention to the effects of pre-training on domain transfer tasks. In this paper, we provide a broad study and in-depth analysis of pre-training for domain adaptation and generalization, namely: network architectures, size, pre-training loss, and datasets. We observe that simply using a state-of-the-art backbone outperforms existing state-of-the-art domain adaptation baselines and set new baselines on Office-Home and DomainNet improving by 10.7\% and 5.5\%.  We hope that this work can provide more insights for future domain transfer research.


\end{abstract}

\section{Introduction}

It is well-known that deep models often perform poorly on out-of-distribution test data~\cite{hoffman2014one}. Domain transfer has been an active research topic for years, aiming to learn more robust feature representations
that generalize from training data (source domains) to novel data distributions (target domains).
There has been significant progress in domain transfer for visual recognition tasks, such as image classification \cite{ganin2017domain}, semantic segmentation~\cite{tsai2019domain} and object detection~\cite{saito2019strong}. 

\begin{figure}[t]
\centering
    \begin{subfigure}[t]{0.49\textwidth}
        \centering
        \includegraphics[width=0.8\linewidth]{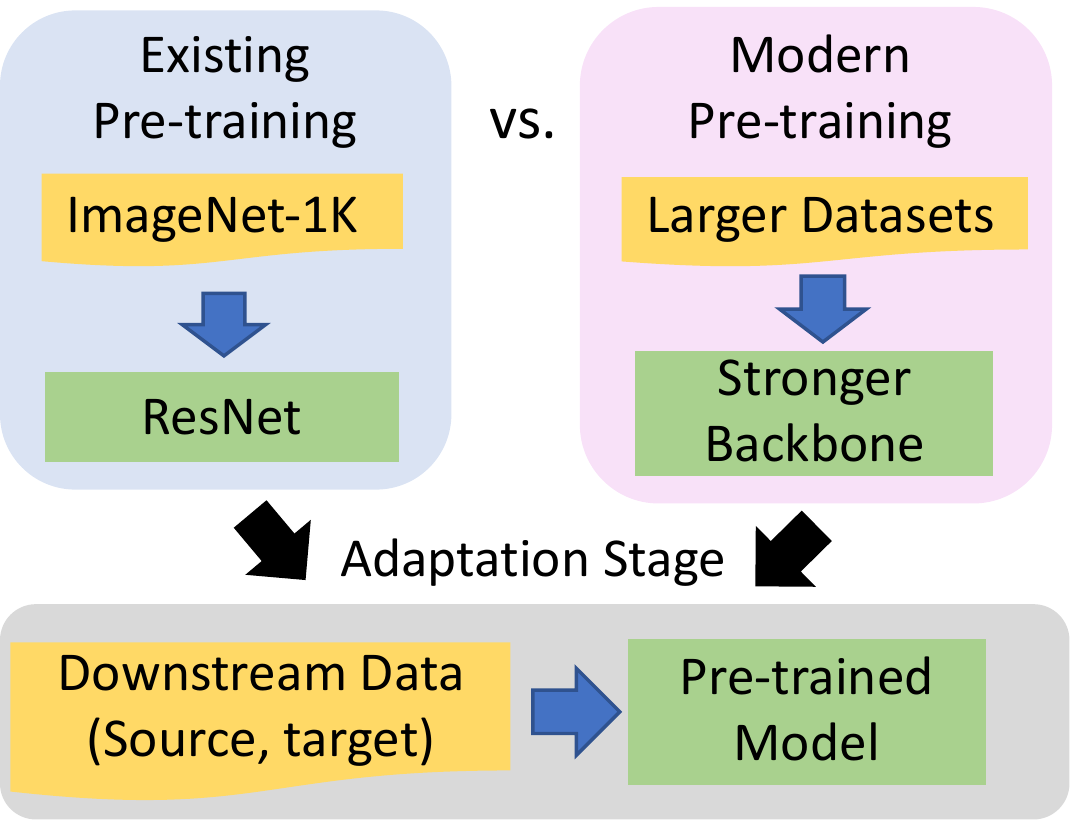}   
        \caption{Two-stage Training}
    \end{subfigure}
    \begin{subfigure}[t]{0.49\textwidth}
        \includegraphics[width=1.0\linewidth]{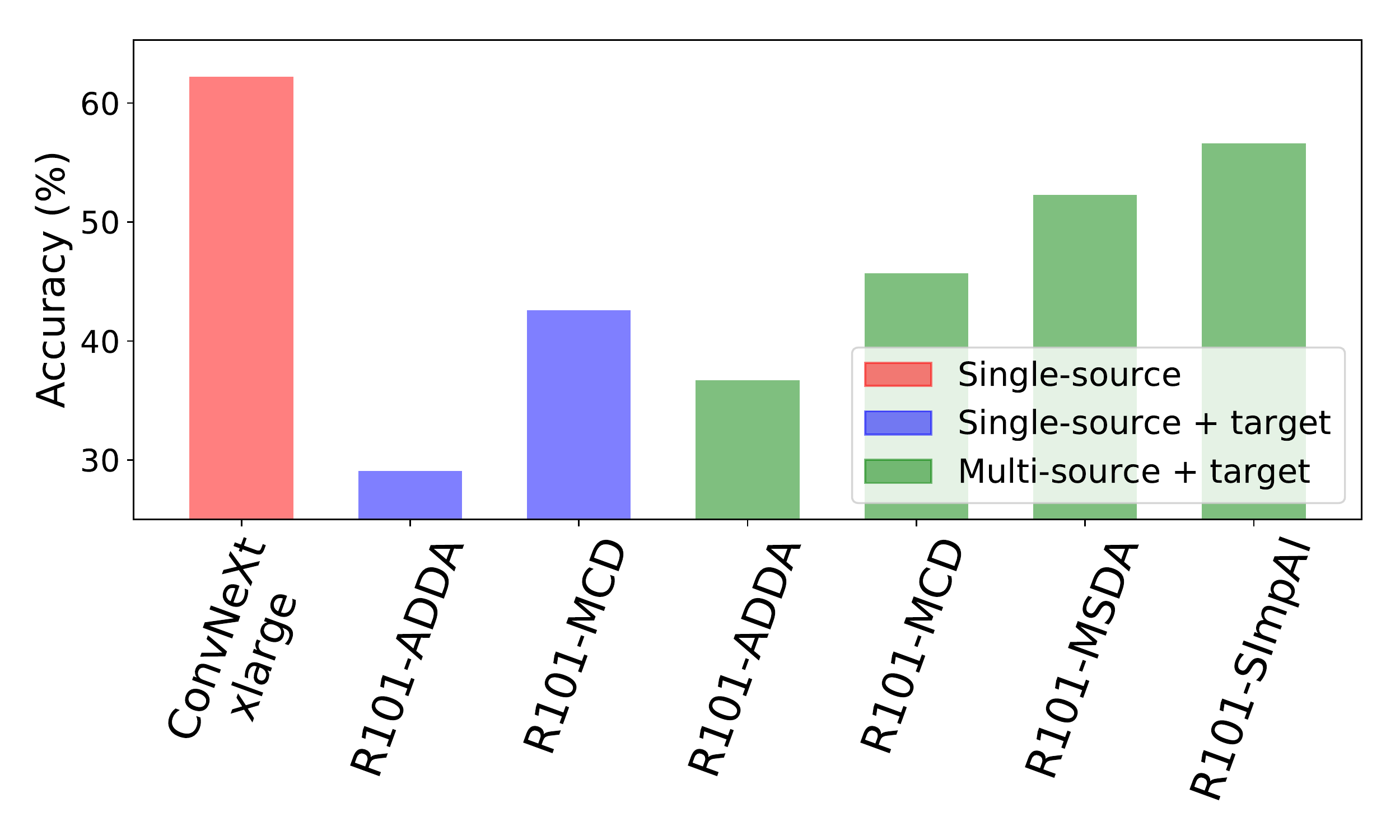}  
        \caption{Target: DomainNet-Painting}
    \end{subfigure}
    \caption{ (a) Existing domain transfer approaches use a ResNet pre-trained on ImageNet-1K and focus on the adaptation stage. We analyze the effect of  pre-training on the adaptation stage. (b) Simply using strong pre-training outperforms all domain adaptation (DA) baselines. We compare accuracy on the target domain (\textit{Painting}) with a SOTA architecture, \textit{ConvNeXt-XL}, pre-trained on ImageNet 22k vs. the standard \textit{ResNet backbones} pre-trained on ImageNet-1K. \textcolor{red}{Red} bars represent the accuracy of model trained only on a single source (\textit{Real} domain). \textcolor{blue}{Blue} bars represent DA models trained on the single source and the unlabeled target domain. \textcolor{green}{Green} bars represent DA models trained on multiple source domains and the unlabeled target domain. ConNeXt-XL fine-tuned only on a single source domain outperforms all the existing DA baselines
    }
    \label{fig:overview}
\end{figure}

Domain transfer consists of two steps: 1) \textit{pre-training}, where a model is first pre-trained on an upstream task with a massive supervised dataset, \eg, ImageNet, and 2) \textit{transfer (adaptation)}, where the model is fine-tuned  on downstream multi-domain data, see Fig.~\ref{fig:overview}-(a). In the latter step,
\textit{Domain Adaptation (DA)} tunes on both a labeled source and an unlabeled target domain, while \textit{Domain Generalization (DG)} tunes only on labeled source data.
While many DA and DG methods (\eg,~adversarial learning \cite{ganin2017domain,long2018conditional,tzeng2014deep}, entropy optimization~\cite{long2018conditional,saito2019semi} or clustering~\cite{huang2019unsupervised}) have been proposed and studied extensively in prior work, little attention has been paid to pre-training for domain transfer. In this paper, we provide  comprehensive experiments and an in-depth analysis of pre-training.

Pre-training is a very successful transfer learning technique for many visual tasks, including domain transfer tasks, as it provides a strong initial representation~\cite{chen2020simple,kornblith2019better}. Pre-training is especially useful when annotations are limited. We decompose pre-training  into three parts: (a) network architecture (backbone), (b) dataset, and (c) loss function. It is a common practice of most domain transfer methods to use a ResNet backbone pre-trained on ImageNet-1K with a supervised loss function (\ie, cross-entropy loss). We argue that this evaluation standard is outdated and ignores the effect of modern large-scale pre-training on domain transfer.
To illustrate the potential impact of pre-training, Fig.~\ref{fig:overview}-(b) shows an experiment that compares the performance of different backbones to SOTA results on the DomainNet~\cite{peng2019moment} DA benchmark. Simply using a recent backbone~\cite{liu2022convnet} pre-trained with ImageNet-22K with no adaptation outperforms existing domain transfer methods. This raises the question, will SOTA DA methods still provide similar gains if applied to the stronger backbones? 

To fully explore these issues, in this paper we pose the following questions: 
\begin{enumerate}

\item \textbf{What is the effect of network architecture?} ResNet-based backbones~\cite{he2016deep} are commonly used in domain generalization~\cite{gulrajani2021in}, single source~\cite{saito2020uni,ganin2017domain,long2018conditional} and multi-source DA~\cite{peng2019moment,venkat2020your}. Since larger and more powerful backbones such as Swin-Transformer~\cite{liu2021swin} or ConvNext~\cite{liu2022convnet} have been recently proposed, we ask whether they may be more robust to domain shift. Transformers were recently shown to be more robust than CNNs to image corruptions and adversarial examples~\cite{bai2021transformers}. We thus conduct an extensive analysis of the impact of network size and architectures, including state-of-the-art Transformers and CNNs on domain transfer tasks.

\item \textbf{What is the effect of pre-training dataset?} Several datasets that are larger than the standard ImageNet-1K could potentially improve transfer: ImageNet-22K~\cite{russakovsky2015imagenet}, JFT-300M~\cite{hinton2015distilling} and Conceptual Captions~\cite{changpinyo2021conceptual,sharma2018conceptual}. These datasets have been very effective for diverse downstream visual tasks (\eg,~\cite{dosovitskiy2021an,li2021align}), but not well explored for domain transfer tasks. 
We therefore study the effect of a wider range of pre-training datasets, including ImageNet-21K, JFT-300M, and language-vision datasets, on domain transfer. 
    
\item \textbf{Supervised vs. Self-supervised Pre-training.} In terms of loss functions, self-supervised learning (\eg,~\cite{caron2021emerging,chen2020simple,wu2018unsupervised}) has obtained powerful performance on diverse visual tasks and often outperforms its supervised counterparts on downstream problems~\cite{bashkirova2021visda,caron2021emerging,chen2020simple}. We therefore compare  self-supervised and supervised pre-training for domain transfer.

\item \textbf{Domain Adaptation with SOTA Pre-training.} Finally, we investigate a fundamental research question in domain transfer: With the help of the state-of-the-art pre-trained models, do we still need sophisticated domain adaptation methods? We explore the applicability of several existing DA methods to our more advanced pre-training setting.
\end{enumerate}

We conduct the study on four standard multi-domain benchmarks. 
While we find that with better pre-training, DA methods still improve performance compared to a source-only trained model, an outdated DA method outperforms  state-of-the-art DA methods. This raises serious fundamental research questions about the current evaluation protocol. 


In summary, our work's main contribution is to provide the field with a broad comparison of modern pre-training approaches for domain transfer tasks. To our knowledge, this is the first work to do such an in-depth analysis. One of our key findings is that SOTA pre-training outperforms SOTA domain transfer methods by a large margin even without access to a target domain, as shown in Fig.~\ref{fig:overview}-(b). We also observe network architectures, sizes, and pre-training datasets play a big role but in a domain-dependent way. Finally, we show that SOTA DA methods work less than older DA methods under modern pre-training. We hope our work will modernize  current domain transfer benchmarks and provide helpful and practical insights for future domain adaptation research.


\section{Related Work}

\noindent\textbf{Domain Transfer.} Domain transfer tasks aim to improve generalization and mitigate domain shift between source and target domains. We study generalization to \textit{natural} data shifts caused by the changes in visual styles, background, lighting, etc.~\cite{hendrycks2018benchmarking,koh2021wilds,saenko2010adapting} as opposed to artificial corruptions~\cite{goodfellow2014explaining}. In this problem setup, we are given a single source domain or multi-source labeled domains. The key is how to learn transferable features that will be useful for the unlabeled target domain. Depending on the specific  setup, this task can be categorized into two: (1) domain adaptation (DA) where we can access the target domain and (2) domain generalization (DG) where we do not have access to a target domain. Depending on the number of labeled source domains, each category can be further divided into single-source or multi-source DA (or DG). Typically, there are two stages: (1) pre-training and (2) adaptation. Most of these methods focus on the second adaptation stage for domain alignment with adversarial domain classifier~\cite{ganin2017domain,long2018conditional}, entropy optimization~\cite{long2018conditional,saito2019semi,kim2021cds}, minimizing maximum discrepancy across domain distributions~\cite{saito2018maximum,zhang2019bridging}, maximum mean discrepancy~\cite{long2016unsupervised}, or optimal transport~\cite{bhushan2018deepjdot}. While domain alignment methods have been proposed actively in recent years for the adaptation stage, the importance of the pre-training stage has not been well explored. Pre-training can provide strong weight initialization by learning a general transferable representation that can be useful for diverse downstream tasks~\cite{kornblith2019better}. While typical DA or DG methods use ResNet backbones pre-trained on ImageNet-1K, we focus on the pre-training stage and provide an in-depth analysis of its effects on domain transfer tasks.

\noindent\textbf{Network Architectures and Datasets for Pre-training.} 
Since the transferability of the model is closely correlated with the performance on downstream tasks as shown in~\cite{kornblith2019better}, having a strong pre-trained model is important. In terms of architectures, convolutional neural networks (CNN) has been standard and state-of-the-art models in many visual tasks for years. After the introduction of AlexNet~\cite{krizhevsky2012imagenet} with ImageNet-1K~\cite{russakovsky2015imagenet}, new CNN-based architectures have been proposed with deeper, wider, and more effective convolutional layers, \eg, VGGNet~\cite{simonyan2014very}, ResNe(X)t~\cite{he2016deep,xie2017aggregated}, SENet~\cite{hu2018squeeze}, EfficientNet~\cite{tan2019efficientnet}, and ConvNeXt~\cite{liu2022convnet}. A newer line of work uses self-attention layers or Transformers for vision. Inspired by the Transformers for NLP~\cite{vaswani2017attention}, transformers for vision have been introduced in Vision Transformers (ViT)~\cite{dosovitskiy2021an} and shows encouraging results by training with larger training sets than ImageNet-1K such as JFT-300M~\cite{sun2017revisiting} or ImageNet-22K. DeiT~\cite{touvron2021training} propose an efficient training strategy to train ViT. Swin Transformers employ a hierarchical transformer with a sliding window strategy where self-attention is performed within a local window. Swin Transformers achieve state-of-the-art performance in a range of computer vision tasks including object detection and segmentation. Bai~\etal~\cite{bai2021transformers} show that Transformers can improve the generalization capability on out-of-distribution samples compared to CNNs. However, Liu~\etal~\cite{liu2022convnet} propose ConvNeXt, which modernizes the ResNet architecture and show that a CNN can still outperform Transformers in vision and more robust to distribution shift. In addition to network architectures, it is shown that larger pre-training datasets such as ImageNet-22K, JFT-300M, or image-text pairs can further improve the transferability~\cite{dosovitskiy2021an,jia2021scaling,li2021align,liu2022convnet,radford2021learning,xie2020self}.   Inspired by these observations, we further study the effect of backbones and pre-training datasets on domain transfer evaluation benchmarks.

\noindent\textbf{Self-supervised learning.} Self-supervised learning~\cite{dosovitskiy2015discriminative,gidaris2018unsupervised,noroozi2016unsupervised,wu2018unsupervised} devises pretext tasks with self-supervisory signals without requiring human annotations. These pretext tasks allow a model to learn discriminative and transferable representations with only unlabeled data for later use in downstream tasks. Representative methods include: solving a jigsaw puzzle~\cite{noroozi2016unsupervised}, rotation prediction~\cite{gidaris2018unsupervised}, Instance Discrimination (ID)~\cite{chen2020simple,he2020momentum,wu2018unsupervised}, contrasting cluster assignments~\cite{caron2020unsupervised}, self knowledge distillation~\cite{caron2021emerging}, and masked image modeling~\cite{bao2022beit}. Instance Discrimination~\cite{wu2018unsupervised} learns an embedding that maps visually similar images closer to each other and far from dissimilar images by classifying an image as its unique class. Some of these self-supervised methods outperform the supervised pre-training on several downstream tasks. For example, SwAV outperforms its supervised pre-training on object detection and image classification tasks on VOC~\cite{van2018inaturalist} and INaturalist~\cite{van2018inaturalist}. It is notable that in the VisDA-2021 competition for universal domain adaptation~\cite{bashkirova2021visda}, the self-supervised masked image modeling approach with a transformer backbone~\cite{bao2022beit} is the first place solution. We further investigate the effect of self-supervised pre-training approaches for domain transfer tasks.


\section{Analysis Setup}

\label{sec:analysis_setup}
Our goal is to analyze the effect of pre-training on domain transfer tasks. 
We assume a single source domain $\mathcal{D}_{s}=\left\{\left(\mathbf{x}_{i}^{s}, y_{i}^{s}\right)\right\}_{i=1}^{N_{s}}$ with $N_s$ images $x$ and labels $y$ 
and an unlabeled target domain $\mathcal{D}_{t}=\left\{\mathbf{x}_{i}^{t}\right\}_{i=1}^{N_{t}}$. Given a pre-trained model $f$, we evaluate two types of domain transfer tasks: 1) domain generalization, \ie~fine-tune $f$ on $\mathcal{D}_{s}$ and test on $\mathcal{D}_{t}$, and 2)  domain adaptation, \ie~fine-tune $f$ on $\mathcal{D}_{s},\mathcal{D}_{t}$ and test on $\mathcal{D}_{t}$.



\noindent \textbf{Pre-training Datasets.} Typically, ImageNet-1K is widely used for pre-training. ImageNet-1K contains 1.2M images of mutually exclusive 1000 classes. ImageNet-22K (the superset of ImageNet-1K) is also used for pre-training (\eg,~\cite{liu2021swin,liu2022convnet}), which contains 14.1M images of 22K classes. In addition, Xie \etal~\cite{xie2020self} use a larger dataset JFT-300M to further improve the accuracy. Recently, language-vision models~\cite{jia2021scaling,radford2021learning,li2021align} can be used in image classification using image and text description pairs. We choose ALBEF~\cite{li2021align}, which achieves the-state-of-the-art performance and uses publicly available language-vision datasets. In total, ALBEF is pre-trained on ImageNet-1K, web crawled datasets (Conceptual
Captions~\cite{changpinyo2021conceptual,sharma2018conceptual}, SBU Captions~\cite{ordonez2011im2text}) and two human annotated datasets (COCO\cite{lin2014microsoft} and Visual Genome~\cite{krishna2017visual}). We explore models pre-trained on these datasets.

\noindent \textbf{Downstream Datasets.} We choose Office-Home (OH)~\cite{venkateswara2017deep}, DomainNet (DN)~\cite{peng2019moment}, CUB~\cite{wang2020progressive,wah2011caltech}, and iWildCAM2020 (WILD)~\cite{beery2020iwildcam,koh2021wilds,sagawa2022extending}. Office-Home contains 15K images from 4 domains (Real (Rw), Painting (Pa), Clipart (Cl), Art (Ar)) on 65 classes. DomainNet contains 586K images from 6 domains (Clipart (Cl), Infograph (In), Painting (Pa), Quickdraw (Qu), Real (Rw), Sketch(Sk)) on 345 classes. Office-Home and DomainNet contain many common classes with ImageNet such as a chair, clock, and table. CUB contains 15K images from two domains (Real and Painting) on 200 fine-grained bird classes. For WILD, the source domain and target domain contains 182 different animal species from camera traps in different locations spread across multiple countries in the world. The source domain contains 129K images from 243 camera traps and the target domain contains 14K from 32 camera traps. Then we use the test set with 42K images from 48 different camera traps. The sets of camera traps in source and target domains are disjoint. The images of WILD are all realistic images from camera traps.  ImageNet contains many animal classes, but the annotations are more fine-grained in CUB and WILD. While the main cause of domain-shift is visual styles in Office-Home, DomainNet, CUB, the domain-shift of WILD is mainly caused by location differences between camera traps in the world. 

\noindent \textbf{Backbone.} For CNNs, we investigate the variants of ResNet~\cite{he2016deep}, EfficientNet~\cite{tan2019efficientnet}, and \convnext~\cite{liu2022convnet}. For Transformers, we explore the variants of \vit~\cite{dosovitskiy2021an}, \deit~\cite{touvron2021training} and \swin~\cite{liu2021swin}. Variants include different depths and sizes (\eg, \swin-\{S,B,L\}).

\noindent \textbf{Self-supervised Learning for Pre-training.} In addition to the supervised pre-training, we also explore recent self-supervised learning approaches for pre-training. We study SwAV~\cite{caron2020unsupervised}, MoCo~\cite{he2020momentum}, DINO~\cite{caron2021emerging}, and BEiT~\cite{bao2022beit}.

\noindent \textbf{Challenges of Evaluating Models.} We use pre-trained models, which are publicly available. One of the big challenges is that it is not possible to fairly compare all possible combinations of pre-training datasets, backbones, and self-supervised learning. For example, SwAV (self-supervised method) only provides pre-trained models on one architecture (\eg, ResNet-50) and ImageNet-1K. Therefore, it is not possible to fairly compare these with self-supervised pre-training and supervised pre-training, which use the state-of-the-art architecture (\eg, ConvNeXt) and larger datasets (\eg, ImageNet-22K).

\noindent \textbf{Fine-tuning Details.} From each pre-trained model, we fine-tune the model with a downstream dataset. We use source domain data to train a model and keep 20\% of the source data as a validation set. We choose the learning rate and optimizer by tuning on the validation set. We test different learning rates (lr=1e-1, 1e-2, 1e-3) of SGD and learning rates (lr=1e-3, 1e-4, 1e-5) of the Adam optimizer. We add a new FC layer for downstream tasks and train it from scratch with a learning rate 10 times that of the pre-trained layers. We use the image size of $224\times224$ with random resized cropping. We also use random color jittering, gray-scaling, and horizontal flipping for augmentation for all models. Additional training details can be found in appendix.



\section{Experiments}

 In this section, our goal is to explore the effects of pre-training for domain transfer. We reiterate that most of the prior domain adaptation (DA) or generalization (DG) work use a ResNet backbone pre-trained on ImageNet-1K. A model is denoted by \textit{X-Y} where X represents the name of architecture and Y represents the size of the backbone. For example, Swin-T, Swin-S, and Swin-B represent the tiny, small, and base model of Swin Transformer.  Unless specified otherwise, pre-trained models are trained with a supervised loss (\ie, cross-entropy loss).  We now evaluate different pre-trained models in domain transfer tasks. In Sec.~\ref{sec:single-source}, we investigate single source DG and analyze the architecture, pre-training datasets, and loss functions. We also compare these models with the existing DA works. In Sec.~\ref{sec:DA}, we explore the existing DA with new architectures. Lastly, we provide feature analysis in Sec.~\ref{sec:feature_analysis}.  Our code is available at: \url{https://github.com/VisionLearningGroup/Benchmark_Domain_Transfer}.

\subsection{Single Source Domain Generalization}
\label{sec:single-source}
For this experiment, we fine-tune different pre-trained models with only a single source domain. We take the $Real$ domain as the source domain on Office-Home, DomainNet, and CUB and treat the remaining domains as target domains. For WILD, we follow the split in~\cite{koh2021wilds}. We do not use the target domain data.
\begin{table}[t]
\caption{ Accuracy comparison on architectures in single source domain generalization. Each backbone pre-trained on ImageNet1-K is fine-tuned on a single domain ($Real$) and tested on the remaining domains in each benchmark. Recent architectures achieve higher accuracy than ResNet}
\setlength{\tabcolsep}{4pt}
\resizebox{\textwidth}{!}{
\begin{tabular}{l|c|c|c|c|c|c|c|c|c|c|c|c|c}
\toprule[1.0pt]

\multirow{2}{*}{Backbone} & \multirow{2}{*}{Pre-train. Data} & \multirow{2}{*}{Params} & \multicolumn{3}{c|}{Office-Home} & CUB & WILD & \multicolumn{5}{c|}{DomainNet} & \multirow{2}{*}{AVG} \\
\cline{4-13}
 &  &  & Ar & Cl & Pr & Pa & - & Cl & In & Pa & Qu & Sk &  \\
\hline
\resnet-50 & ImageNet-1K & 23 M & 66.1 & 49.0 & 77.2 & 42.3 & 70.7 & 46.6 & 17.3 & 45.2 & 6.5 & 35.3 & 45.6 \\
\convnext-T & ImageNet-1K & 27 M& 67.4 & 48.7 & 77.9 & 42.5 & 74.0 & \textbf{52.8} & \textbf{20.6} & \textbf{50.8} & \textbf{7.8} & \textbf{41.2} & 48.4 \\
\deit-S & ImageNet-1K & 21 M& 70.0 & \textbf{51.3} & \textbf{81.2} & \textbf{58.0} & 73.4 & 49.5 & 19.4 & 49.2 & 6.9 & 36.1 & \textbf{49.5} \\
\swin-T & ImageNet-1K & 27 M& \textbf{71.3} & 49.4 & 81.1 & 52.0 & \textbf{74.2} & 51.8 & 19.8 & 49.3 & 7.3 & 37.3 & 49.3 \\
\hline
\hline
\resnet-101 & ImageNet-1K & 42 M& 68.5 & 52.4 & 79.9 & 46.1 & 74.0 & 49.3 & 19.2 & 48.6 & \textbf{8.7} & 38.5 & 48.5 \\
\convnext-S & ImageNet-1K & 49 M&72.2& 52.7	&	80.9	&43.7&	76.2&	54.9&	22.2&	\textbf{52.8} &	8.1&	\textbf{43.0}&	50.7 \\
\swin-S & ImageNet-1K & 48 M & \textbf{73.8} & \textbf{54.5} & \textbf{84.2} & \textbf{56.5} & \textbf{78.6} & \textbf{55.9} & \textbf{22.5} & 51.8 & 8.6 & 41.4 & \textbf{52.8} \\

\bottomrule[1.0pt]
\end{tabular}}
\label{tab:single_source}
\end{table}

\noindent\textbf{Analysis of Network Architectures.}
We first compare generalization performance of architectures in Table~\ref{tab:single_source}. All models are pre-trained on ImageNet-1K.  \convnext~ and Transformers (\deit, \swin) outperform their \resnet~ counterparts. In this experiment, Transformer models achieve the highest accuracy on average. \swin-T outperforms ResNet-50 by 3.7\%. This improvement becomes larger in the deeper model. \swin-S outperforms \resnet-101 by 4.3\%. We further analyze the effect of depth in a later section. The big difference between \convnext~ and \trans~ is in the CUB experiment. While \deit-S~significantly improves the accuracy by 15.7\%, \convnext-T~ could not improve much compared to \resnet. However, \convnext~ attains slightly higher accuracy on DomainNet compared to \swin. This suggests that CNN and \trans ~may be robust to different types of domain shift. We put additional results of larger networks (\eg, \deit-B) in the appendix.

\begin{figure}[t!]
\centering
    \includegraphics[width=1.0\linewidth]{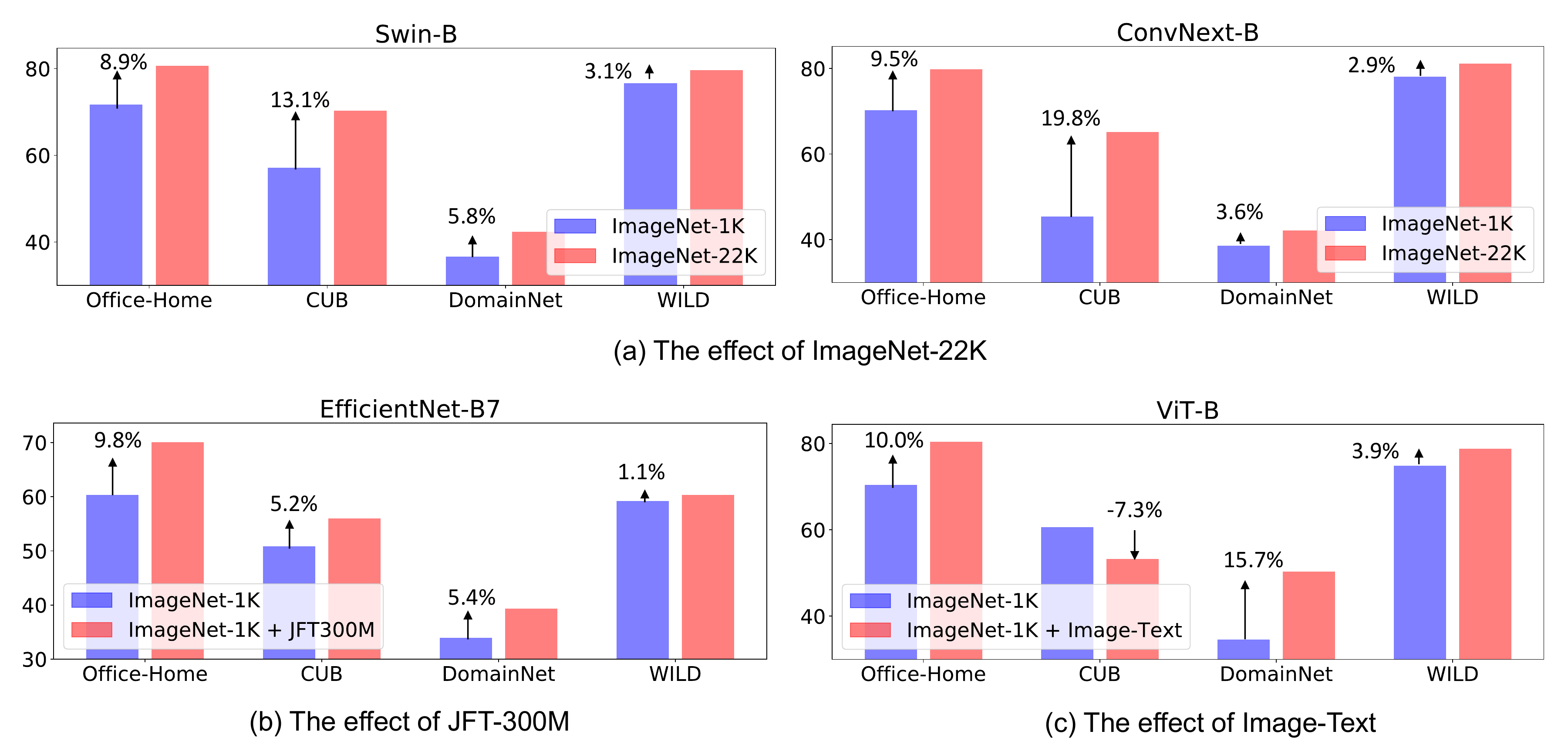}   
    \caption{The effect of additional datasets in pre-training on single source domain generalization. We investigate the following pre-training datasets, (a) ImageNet-22K, (b) JFT-300M, and (c) Image-Text pairs }
    \label{fig:dataset}
\end{figure}

\noindent\textbf{Analysis of Pre-training Datasets.}
We now analyze the effect of additional datasets during pre-training as shown in Fig.~\ref{fig:dataset}. Due to the availability of pre-trained models (see \textit{Challenges  of  Evaluating  Models} in the above section), we use different backbones to compare the effect of pre-training datasets. In Fig.~\ref{fig:dataset}-(a), we compare the accuracy between ImageNet-1K and ImageNet-22K on \swin-B and \convnext-B for Office-Home and DomainNet. We report the accuracy averaged over all settings in each benchmark. In both architectures, pre-training with ImageNet-22K boosts the accuracy for all benchmarks. Especially, there are significant boosts in accuracy on Office-Home and CUB. To be specific, the accuracy of \convnext-B on CUB increases by 19.8\%. In Fig.~\ref{fig:dataset}-(b), we study the effect of JFT-300M. Since a supervised pre-trained model on JFT-300M is not released publicly, we use the publicly available self-trained EfficientNet-B7~\cite{xie2020self} on ImageNet-1K and JFT-300M. While it shows similar accuracy improvements on Office-Home, the accuracy improvements on CUB are smaller than ImageNet-22K. This could be because  self-training uses pseudo-labeling and the number of classes is still limited to  1K. In Fig.~\ref{fig:dataset}-(c), we study the effect of vision-language datasets containing image-text pairs. We study ALBEF, which uses \vit-B as an image encoder and BERT$_{base}$ as a text encoder. We made one modification for ALBEF. We first extract the sentence representation from the text encoder with the prompt template ``A photo of a \{label\}'' following~\cite{radford2021learning}. Then we initialize the weights of the last FC layer with the sentence representations. While it shows  similar behavior on Office-Home, it seriously hurts the performance of CUB. In contrast, ALBEF obtains the highest accuracy on DomainNet compared to all the other models. This indicates that improvement depends on both the pre-training dataset and the downstream task at the same time.

\begin{figure}[t!]
\centering
    \includegraphics[width=0.85\linewidth]{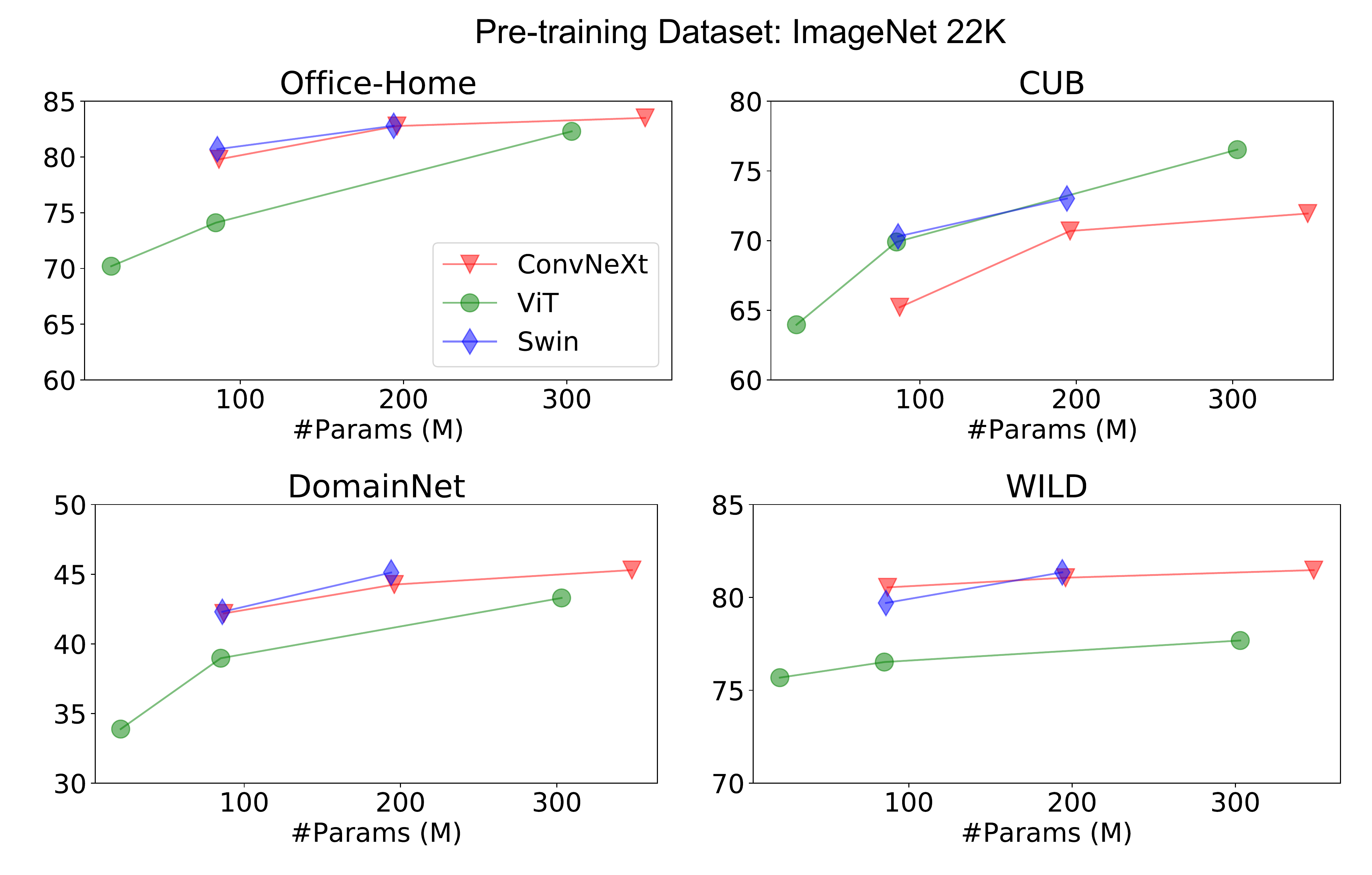}   
    \caption{The effect of architecture depth (number of layers) on single source domain generalization. While \convnext~and \swin~ tend to have slight improvements in accuracy, but the accuracy of \vit~is more sensitive to its depth}
    \label{fig:network_depth}
\end{figure}

\noindent\textbf{Analysis of Network Depth.}
We investigate the accuracy gained from the deeper layers. In Fig.~\ref{fig:network_depth}, we report the accuracy of \convnext-\{B,L,XL\}, \swin-\{B,L\}, and \vit-\{S,B,L\} pre-trained on ImageNet-22K. In general, \vit~ shows bigger changes in accuracy according to its depth. The accuracy of \swin~ and \convnext~ increases slightly on the benchmarks except for CUB. In WILD, all the architectures show minimal changes according to their depth. When comparing the adaptation methods, we often choose a shallow and light model to reduce computational costs. For example, most of the domain generalization methods~\cite{gulrajani2021in} employ ResNet-18 rather than ResNet-101. However, the results suggest that we should be careful when choosing a backbone. If we compare adaptation methods and want to use a shallow model for efficiency, it is more desirable to use \swin~or \convnext~than \vit.


\begin{SCtable}[][t!]
\setlength{\tabcolsep}{4pt}
\resizebox{0.5\textwidth}{!}{\begin{tabular}{l|c|c|c|c|c}
\toprule[1.0pt]
\multirow{2}{*}{Backbone} & \multirow{2}{*}{Training} & \multicolumn{4}{c}{Dataset} \\
\cline{3-6}
 &  & OH & CUB & WILD & DN \\
 \hline
 \hline
\multicolumn{6}{l}{\textbf{(a) Pre-training Data: ImageNet-1K}} \\
\resnet-50 & Sup. & \textbf{64.1} & \textbf{42.3} & \textbf{70.7} & 30.2 \\
\resnet-50 & MoCo & 53.7 & 37.1 & 67.9 & 28.0 \\
\resnet-50 & SwAV & 56.5 & 39.2 & 68.4 & \textbf{30.7} \\
\resnet-50 & DINO & 56.6 & 37.2 & 67.6 & 23.1 \\
\hline
\deit-S & Sup. & \textbf{58.0} &	\textbf{58.0} & 73.4 &	\textbf{32.2} \\
\vit-S & DINO & 56.3 & 57.9 & \textbf{75.1} & 29.8 \\
\hline
\hline
\multicolumn{6}{l}{\textbf{(b) Pre-training Data: ImageNet-22K}} \\
\vit-L & Sup. & \textbf{82.3} & 76.5 & 77.7 & 43.3 \\
\vit-L & Sup.+BEiT & 80.9  & \textbf{78.4}  & \textbf{79.7} & \textbf{45.1} \\
\bottomrule[1.0pt]
\end{tabular}}
\caption{ Accuracy comparison between supervised and self-supervised learning in pre-training. (a) In this domain generalization task, supervised pre-trainining outperforms the self-supervised pre-training in most cases. (b) When self-supervised learning is combined with supervised learning, the performance can be improved}
\label{tab:self_sup}
\end{SCtable}

\noindent\textbf{Supervised vs. Self-supervised Learning.}
Recently, several self-supervised learning (\eg,~\cite{caron2021emerging,chen2020simple,wu2018unsupervised}) methods outperformed their supervised counterparts on various downstream tasks~\cite{bao2022beit,bashkirova2021visda,caron2020unsupervised,caron2021emerging,chen2020simple}. In Table~\ref{tab:self_sup}, we investigate the effect of self-supervised learning approaches in pre-training for domain transfer. In Table~\ref{tab:self_sup}-(a), we observe that supervised learning (denoted as Sup.) performs better than self-supervised learning in most cases. Especially on Office-Home and CUB, self-supervised learning significantly hurts performance compared to supervised learning. In Table~\ref{tab:self_sup}-(b), we explore the \beit, the winner of the VisDA 2021 DA competition~\cite{bashkirova2021visda}. \beit~uses both self-supervised and supervised learning. When self-supervised learning is combined with supervised learning and improves the performance on CUB, DomainNet (denoted as DN), and WILD. While most of the self-supervised approaches focus on only unlabeled data, combining it with labels should be considered to improve further in future research.

\begin{table}[t!]
\caption{Accuracy comparison with domain adaptation baselines on (a) Office-Home and (b) DomainNet. (\rom{1}): For single source adaptation, an unlabeled target domain is used for alignment with a single source domain. (\rom{2}) For multi source adaptation, all the domains except the target domain in each benchmark are used as the source domains along with the unlabeled target domain. (\rom{3}) We change the pre-training with the state-of-the-art backbone and larger datasets. Even though these are trained with only one single domain \textbf{without any adaptation}, simply using the state-of-the-art pre-training outperforms the existing domain adaptation baselines by up to 10.7\% and 5.5\% on average on each dataset. *~represents the numbers reported in~\cite{jin2020minimum,peng2019moment,tang2020unsupervised,venkat2020your}}
\centering
\setlength{\tabcolsep}{4pt}

\begin{subtable}[t!]{1.0\textwidth}
\caption{Office-Home}
\resizebox{1.0\textwidth}{!}{\begin{tabular}{l|c|c|c|c|c|c|c}
\toprule[1.0pt]
 \multirow{2}{*}{Backbone} & Pre-training  & Downstream  & \multirow{2}{*}{Adaptation} & \multicolumn{3}{c|}{Target Domain} & \multirow{2}{*}{AVG} \\
 \cline{5-7}
 & Data & Data &  & Ar & Cl & Pr &  \\
 \hline
 \hline
\multicolumn{8}{l}{\textbf{(\rom{1}) Single source domain adaptation baselines}} \\
ResNet-50* & ImageNet-1K & Source + Target & DANN~\cite{ganin2017domain} & 63.2 & 51.8 & 76.8 & 63.9 \\
ResNet-50* & ImageNet-1K & Source + Target & CDAN~\cite{long2018conditional} & 70.9 & 56.7 & 81.6 & 69.7 \\
ResNet-50* & ImageNet-1K & Source + Target & AFN~\cite{xu2019larger} & 70.9 & 57.1 & 81.5 & 69.8 \\
ResNet-50* & ImageNet-1K & Source + Target & MDD~\cite{zhang2019bridging} & 72.5 & 60.2 & 82.3 & 71.7 \\
ResNet-50* & ImageNet-1K & Source + Target & SRDC~\cite{tang2020unsupervised} & 76.3 & 57.1 & 85.0 & 72.8 \\
\hline
\multicolumn{8}{l}{\textbf{(\rom{2}) Multi source domain adaptation baselines}} \\
ResNet-50* & ImageNet-1K & Multi source + Target & MFSAN~\cite{zhu2019aligning} & 72.1 & 62 & 80.3 & 71.5 \\
ResNet-50* & ImageNet-1K & Multi source + Target & SImpAl~\cite{venkat2020your} & 70.8 & 56.3 & 80.2 & 69.1 \\
ResNet-101* & ImageNet-1K & Multi source + Target & SImpAl~\cite{venkat2020your} & 73.4 & 62.4 & 81.0 & 72.3 \\
\hline
\multicolumn{8}{l}{\textbf{(\rom{3}) Single source only baselines without adaptation}} \\
ResNet-50* & ImageNet-1K & Source & \xmark  & 53.9 & 41.2 & 59.9 & 46.1 \\
\convnext-XL & ImageNet-22K & Source & \xmark & \textbf{85.1} & 74.0 & \textbf{91.4} & \textbf{83.5} \\
\vit-L & ImageNet-22K & Source & \xmark & 84.0 & 73.0 & 89.9 & 82.3 \\
\swin-L & ImageNet-22K & Source & \xmark & 83.4 & \textbf{74.3} & 90.9 & 82.8 \\
\vit-B & ALBEF~\cite{li2021align} & Source & \xmark & 81.7  & 72.5& 87.2 & 80.4 \\
\bottomrule[1.0pt]
\end{tabular}}
\end{subtable}

\begin{subtable}[t!]{1.0\textwidth}
\caption{DomainNet}
\resizebox{1.0\textwidth}{!}{\begin{tabular}{l|c|c|c|c|c|c|c|c|c}
\toprule[1.0pt]
\multicolumn{1}{c|}{\multirow{2}{*}{Backbone}} & \multicolumn{1}{c|}{Pre-training} & \multicolumn{1}{c|}{Downstream} & \multirow{2}{*}{Adaptation} & \multicolumn{5}{c|}{Target Domain} & \multirow{2}{*}{AVG} \\
\cline{5-9}
\multicolumn{1}{c|}{} & \multicolumn{1}{c|}{Data} & \multicolumn{1}{c|}{Data} &  & Cl & In & Pa & Qu & Sk &  \\
\hline
\hline
\multicolumn{10}{l}{\textbf{(\rom{1}) Single source domain adaptation baselines}} \\
ResNet-101* & ImageNet-1K & Source + target & ADDA~\cite{tzeng2014deep} & 39.5 & 14.5 & 29.1 & 14.9 & 30.7 & 25.7 \\
ResNet-101* & ImageNet-1K & Source + target & MCD~\cite{saito2018maximum} & 42.6 & 19.6 & 42.6 & 3.8 & 30.8 & 27.9 \\
\hline
\multicolumn{10}{l}{\textbf{(\rom{2}) Multi source domain adaptation baselines}} \\
ResNet-101* & ImageNet-1K & Multi source + target & ADDA~\cite{tzeng2014deep} & 47.5 & 11.4 & 36.7 & 14.7 & 33.5 & 28.8 \\
ResNet-101* & ImageNet-1K & Multi source + target & MCD~\cite{saito2018maximum} & 54.3 & 22.1 & 45.7 & 7.6 & 43.5 & 34.6 \\
ResNet-101* & ImageNet-1K & Multi source + target & DCTN~\cite{xu2018deep} & 48.6 & 23.5 & 48.8 & 7.2 & 47.3 & 35.1 \\
ResNet-101* & ImageNet-1K & Multi source + target & MSDA~\cite{peng2019moment} & 58.6 & 26.0 & 52.3 & 6.3 & 49.5 & 38.5 \\
ResNet-101* & ImageNet-1K & Multi source + target & SImpAI~\cite{venkat2020your} & 66.4 & 26.5 & 56.6 & 18.9 & 55.5 & 44.8 \\
\hline
\multicolumn{10}{l}{\textbf{(\rom{3}) Single source only baselines without adaptation}} \\
ResNet-101* & ImageNet-1K & Source & \xmark & 48.4 & 22.2 & 49.4 & 6.4 & 38.8 & 33.0 \\
ConvNext-XL & ImageNet-22K & Source & \xmark & 67.7 & 29.7 & 62.2 & 11.4 & 55.5 & 45.3 \\
ViT-L & ImageNet-22K & Source & \xmark & 65.5 & 27.3 & 61.3 & 10.2 & 52.1 & 43.3 \\
Swin-L & ImageNet-22K & Source & \xmark & 67.2 & 30.6 & 62.5 & 11.2 & 54.1 & 45.1 \\
ViT-B & ALBEF~\cite{li2021align} & Source & \xmark & \textbf{73.6} & \textbf{37.3} & \textbf{65.3} & \textbf{12.8} & \textbf{62.2} &\textbf{50.3} \\  
\bottomrule[1.0pt]
\end{tabular}}
\end{subtable}

\label{tab:da_comp}
\end{table}

\noindent\textbf{Comparison with Domain Adaptation Baselines.}
We provide a performance comparison with the existing DA baselines in Table~\ref{tab:da_comp}. In this comparison we use Office-Home and DomainNet, which are most extensively explored in prior work. In each table, (\rom{1}) reports the performance of single source DA, where a labeled source domain and unlabeled target domain data is used together for adaptation. In (\rom{2}), multiple labeled source domains are used in addition to the unlabeled target domain. (\rom{1},\rom{2}) use ResNet backbones pre-trained on ImageNet-1K. (\rom{3}) only uses one single source ($Real$) domain with the recent backbones and larger pre-training datasets. It is surprising that (\rom{3}), the state-of-art backbones pre-trained on a larger dataset, significantly outperform DA baselines (\rom{1}, \rom{2}) despite being trained only on a single source and \textit{not using any adaptation on the target domain}. While ConvNext-XL obtains the best results and outperforms adaptation baselines by up to 10.7\% on Office-Home, ViT-B with ALBEF gains more improvements and outperforms adaptation baselines by up to 5.5\% on DomainNet. This observation raises a question, is it still fair and reasonable to use ResNet backbones pre-trained on ImageNet-1K as a standard backbone for the comparison of adaptation methods? From the results, it is clear that the the the standard backbone in the existing domain adaptation benchmarks are outdated. The pre-training stage for DA needs to be updated to reflect recent advances in pre-training.

\begin{table}[t]
\caption{Single source domain adaptation with the state-of-the-art pre-training. We report the target accuracy on each bencmark. For Office-Home, CUB, and DomainNet, we use the $Real$ domain as a source domain. Surprisingly, more recent adaptation methods (AFN, MDD, MCC) underperform CDAN on average }
\setlength{\tabcolsep}{4pt}
\resizebox{\textwidth}{!}{
\begin{tabular}{l|c|c|c|c|c|c|c|c|c|c|c|c|c}
\toprule[1.0pt]
\multirow{2}{*}{Backbone} & \multirow{2}{*}{Adaptation} & \multicolumn{4}{c|}{Office-Home} & CUB & WILD & \multicolumn{6}{c}{DomainNet} \\
\cline{3-14}
 &  & Ar & Cl & Pr & AVG & Pa & - & Cl & In & Pa & Qu & Sk & AVG \\
 \hline
 \hline
Swin-L & Source-only & 74.3 & 83.4 & 90.9 & 82.8 & 73.0 & 81.4 & 67.2 & 30.6 & 62.5 & 11.2 & 54.1 & 45.1 \\
\convnext-XL & Source-only & 74.0 & \textbf{85.1} & 91.4 & 83.5 & 71.9 & 81.5 & 67.7 & 29.7 &  62.2 &11.4 & 55.5 & 45.3 \\
\hline
Swin-L & DANN~\cite{ganin2017domain} & 87.3 & 79.5 & 93.0 & 86.6 & 82.0 & 68.4 & 70.4 & 36.7 & 66.6 & 13.0 & 60.5 & 49.4 \\
ConvNext-XL & DANN~\cite{ganin2017domain} & 87.2 & 79.8 & 93.1 & 86.7 & 80.4 & 66.2 & 70.2 & 36.7 & 66.6 & 8.6 & 62.1 & 48.8 \\
\hline
 Swin-L& CDAN~\cite{long2016unsupervised} & 90.1 & 81.9 & 93.1 & 88.4 & \textbf{84.3} & 81.5 & 72.0 & \textbf{39.7 }& \textbf{67.5} & 10.8 & 61.8 & 50.4 \\
ConvNext-XL & CDAN~\cite{long2016unsupervised}  & \textbf{90.2} & 84.6 & 93.8 & \textbf{89.5} & 83.7 & \textbf{82.5} &\textbf{ 72.1} & 39.5 & 67.3 & \textbf{13.9} & \textbf{63.4} & \textbf{51.2} \\
\hline
 Swin-L & AFN~\cite{xu2019larger} & 87.4 & 77.6 & 92.0 & 85.7 & 79.9 & 78.9 & 68.0 & 34.5 & 64.7 & 6.9 & 58.1 & 46.4 \\
ConvNext-XL & AFN~\cite{xu2019larger}  & 86.0 & 77.7 & 92.8 & 85.5 & 77.7 & 79.7 & 68.0 & 33.4 & 65.0 & 7.8 & 59.2 & 46.7 \\
\hline
Swin-L  & MDD~\cite{zhang2019bridging} & 87.8 & 78.0 & 93.6 & 86.5 & 83.6 & 81.1 & 62.8 & 34.5 & 60.7 & 2.8 & 46.6 & 41.5 \\
ConvNext-XL & MDD~\cite{zhang2019bridging}  & 88.0 & 77.6 & 93.6 & 86.4 & 78.8 & \textbf{82.5} & 58.9 & 29.6 & 62.8 & 11.4 & 51.4 & 42.8 \\
\hline
Swin-L & MCC~\cite{jin2020minimum} & 89.6 & 81.1 & 94.1 & 88.3 & 66.7 & 79.5 & 71.5 & 36.4 & 66.3 & 3.0 & 58.2 & 47.1 \\
ConvNext-XL  & MCC~\cite{jin2020minimum} & 89.5 & 82.9 & \textbf{94.4} & 88.9 & 78.8 & 79.8 & 71.3 & 27.7 & 66.8 & 1.9 & 60.2 & 45.6 \\
\bottomrule[1.0pt]
\end{tabular}}
\label{tab:all_da}
\end{table}

\subsection{Domain Adaptation with Modern Pre-training}
\label{sec:DA}
The observation in Sec.~\ref{sec:single-source} leads us to the next question, will SOTA DA methods still provide gains when these are applied to the recent stronger architectures and pre-training? We study the transferability of prior adaptation methods with new architectures pre-trained on larger datasets.  

\noindent\textbf{Transferability of Domain Adaptation.}  We employ DANN (JMLR'16), CDAN (NeurIPS'18), AFN (ICCV'19), MDD (ICML'19), and MCC(ECCV'20). Table~\ref{tab:all_da} provides the performance of domain adaptation between \swin-L and ConvNeXt-XL pre-trained on ImageNet-22K. First, DA methods still improve the accuracy on average compared to source only (SO) models. However, we observe negative transfer in some settings where DA hurts the performances compared SO. To be specific, all the adaptation methods badly affect the performance on $Real\rightarrow Clipart (Cl)$ in Office-Home.  Second, the relative ranking among adaptation methods on new architectures is different from the ranking on ResNet-based backboes. While MCC obtains SOTA accuracy on ResNet, but CDAN outperforms AFN, MDD, and MCC with these new architectures, which was proposed earlier than the others. This behavior raises another practical question, which adaptation method should we consider the state-of-the-art? We certainly want to have a model with strong performance, but the existing adaptation benchmark with outdated pre-training can not choose the strongest model. We argue that adaptation methods should have transferability on various backbones in order to avoid the potential risk of overfitting to a specific backbone, so that it is able to successfully transfer to new architectures in future.

\begin{figure}[t]
\centering
    \includegraphics[width=1.0\linewidth]{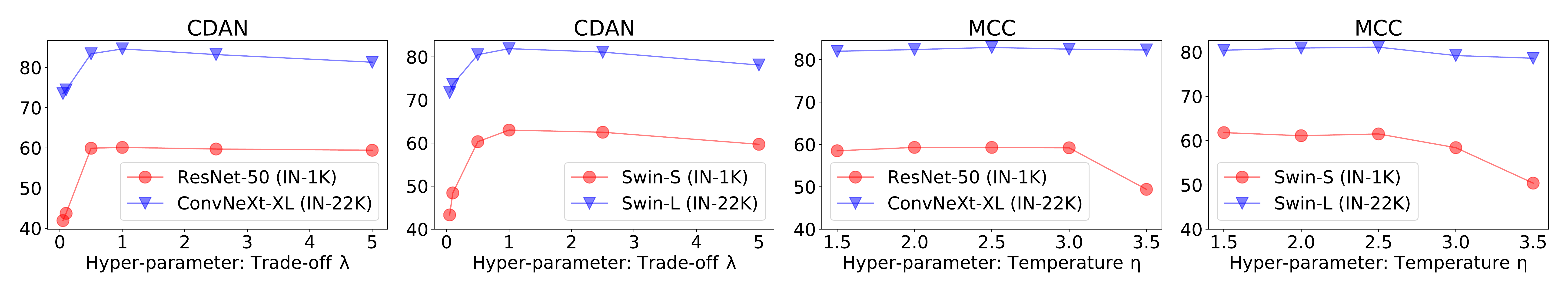}   
    \caption{ Analysis on the transferability of hyper-parameters in adaptation methods between a shallow and deep network. We report the accuracy of CNN and \trans~on the $Real\rightarrow Clipart$ setting in Office-Home. We observe that the shallow and deep network achieve the highest accuracy on the same value of hyper-parameters in both CDAN and MCC}
    \label{fig:hyper_param}
\end{figure}

\noindent\textbf{Analysis on Hyper-parameter.} 
In this experiment, we analyze the transferability of hyper-parameters to new architectures in each adaptation method. The question is whether the optimal hyper-parameters in a shallow network (\eg, ResNet-50) are still optimal in a deep network (\eg, \convnext-XL). We believe this is a practically important question as training a big network is computationally expensive. Hyper-parameter search in a big network can be prohibitively expensive in terms of computational cost. Therefore, the desired property is that hyper-parameters in adaptation methods are transferable between different architectures and their depth. Fig.~\ref{fig:hyper_param} shows the analysis of performance depending on the hyper-parameters between a shallow and deep network. We investigate two adaptation methods CDAN and MCC. Following~\cite{saito2021tune}, we vary the trade-off ($\lambda$ = 0.05, 0.1, 0.5, 1.0, 2.5, 5.0) hyper-parameter in CDAN, which controls the trade-off between domain confusion loss and classification loss on the source domain. The default $\lambda$ in ResNet is 1.0. For MCC, we vary the temperature hyper-parameter ($\eta$=1.5, 2.0, 2.5, 3.0, 3.5), which affects of classifier's confusion loss.  The default $\eta$ in ResNet is 2.5. We employ ResNet-50 as a shallow network and \convnext-XL as a deep network for CNN, and \swin-S as a shallow network and \swin-L as a deep network. The shallow networks are pre-trained on ImageNet-1K and deep networks are pre-trained on ImageNet-22K. The accuracy across the depth (shallow vs. deep) and architectures (CNN vs. \trans) show similar tendency and obtain the highest accuracy with the same hyper-parameter values. Therefore, we observe that the default hyper-parameters of $\lambda$ and $\eta$ in shallow networks are transferable to deep networks. Additionally, the sensitivity of deep networks is small compared to that of shallow networks. We measure the standard deviation of accuracy. \swin-L obtains a standard deviation of $4.2\%$, but \swin-S~obtains a much higher standard deviation of $8.3\%$.

\begin{figure}[t]
\centering
    \begin{subfigure}[t]{0.24\textwidth}
        \centering
        \includegraphics[width=1.0\linewidth]{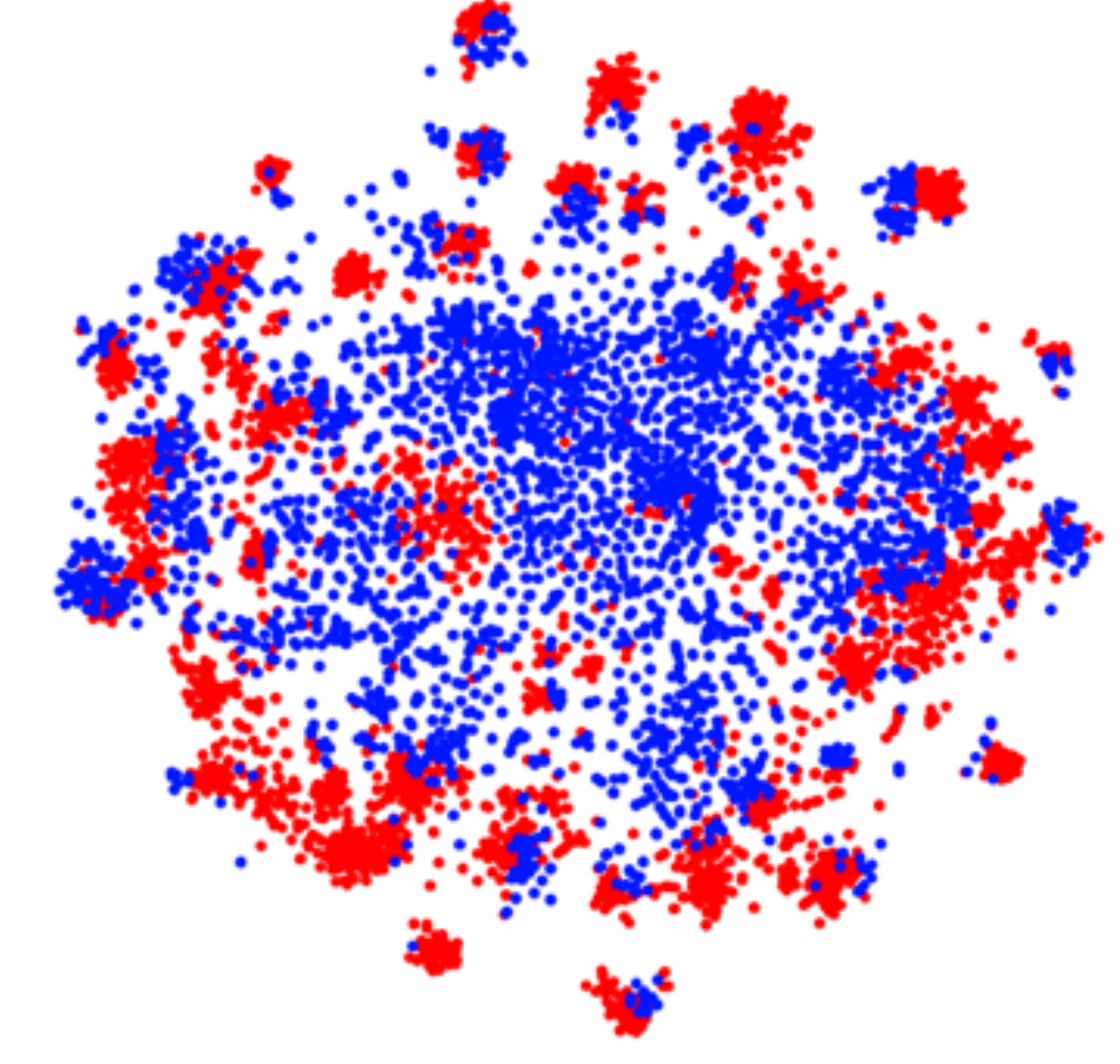}   
        \caption{ResNet-50}
    \end{subfigure}
    \begin{subfigure}[t]{0.24\textwidth}
        \includegraphics[width=1.0\linewidth]{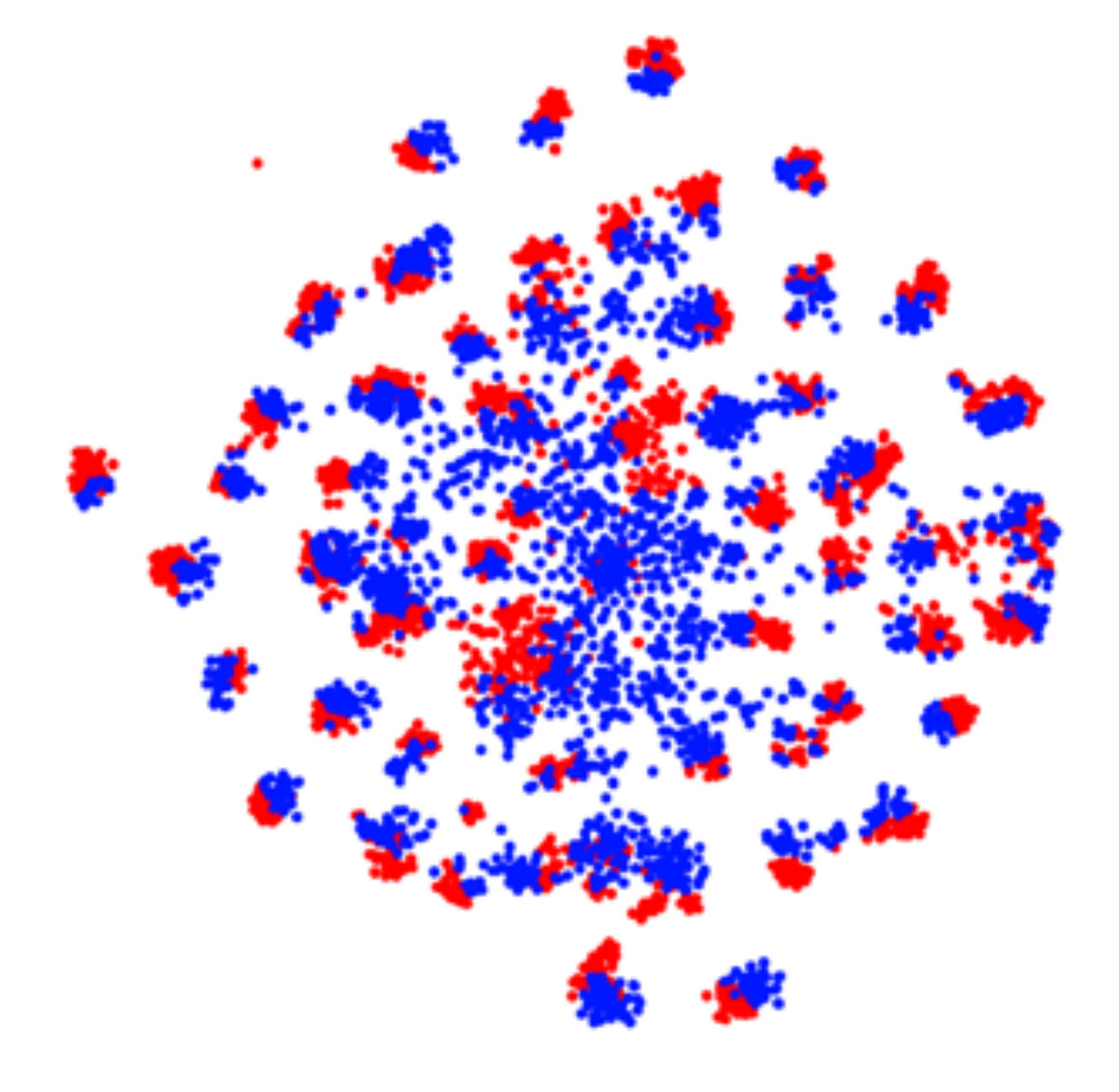}  
        \caption{ConvNext-XL}
    \end{subfigure}
    \begin{subfigure}[t]{0.24\textwidth}
        \includegraphics[width=1.0\linewidth]{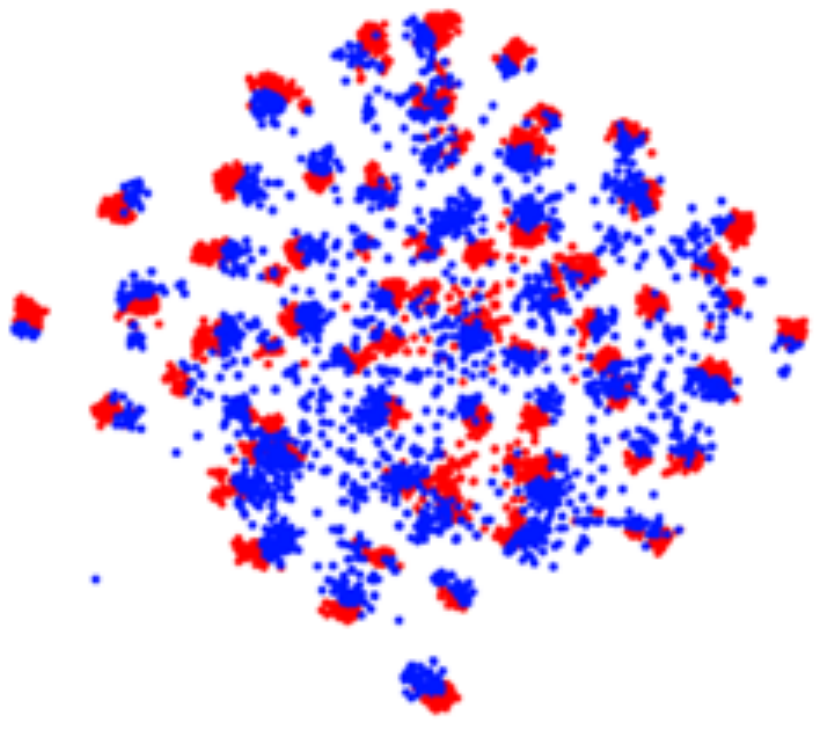}  
        \caption{Swin-L}
    \end{subfigure}
    \begin{subfigure}[t]{0.24\textwidth}
        \includegraphics[width=1.0\linewidth]{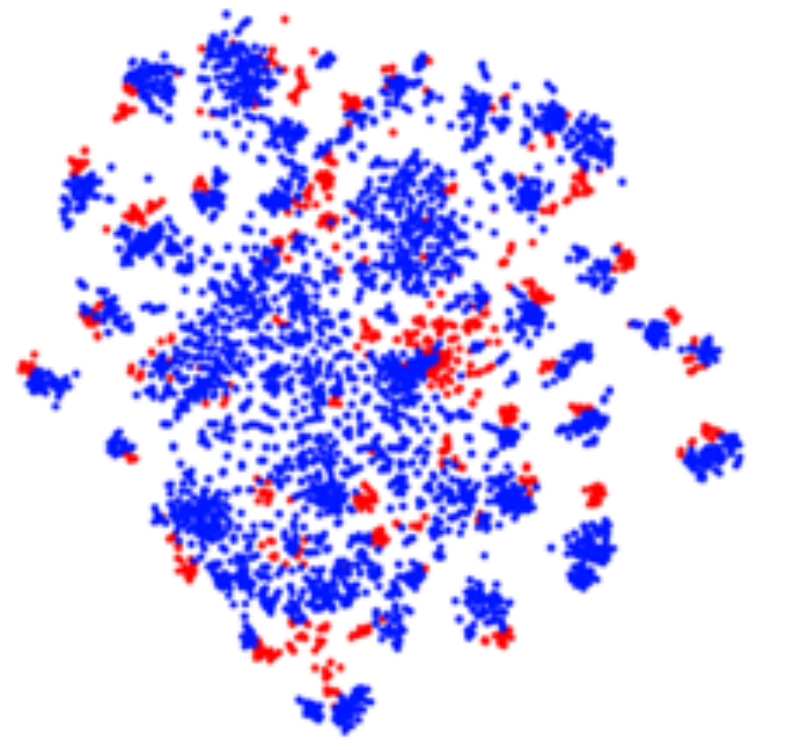}  
        \caption{ViT-B (ALBEF)}
    \end{subfigure}
    \caption{ t-SNE visualization of pre-trained models. We use the Real \textcolor{red}{\textbf{(red)}} and Clipart \textcolor{blue}{\textbf{(blue)}} domains in Office-Home. We directly utilize features from each pre-trained model without fine-tuning. Compared to (a) the ResNet-50 pre-trained on ImageNet-1K, (b) ConvNext-XL and (c) Swin-L pre-trained on ImageNet-22K produce more discriminative as well as domain-aligned representations
    }
    \label{fig:tsne}
\end{figure}


\subsection{Feature Analysis}
\label{sec:feature_analysis}
 We provide feature analysis in this section. We use features directly obtained from each pre-training without fine-tuning on domain adaptation benchmarks. 
 
\noindent\textbf{Feature Visualization.} First, we show the t-SNE~\cite{van2008visualizing} feature visualization of each pre-trained models in Fig.~\ref{fig:tsne}. We compare (a) ResNet-50 pre-trained on ImageNet-1K with (b) \convnext-XL, (c) \swin-L pre-trained on ImageNet-22K, and (d) \vit-B from ALBEF~\cite{li2021align}.  We directly extract features from the pre-trained models on the Real (colored by \textcolor{red}{\textbf{red}}) and Clipart  (colored by \textcolor{blue}{\textbf{blue}}) domains in Office-Home. While the red and blue dots are highly separated in ResNet-50, these are aligned with each other in \convnext-XL and \swin-L. It is also clear that \convnext-XL and \swin-L obtain better clustered and discriminative representations. (d) \vit-B pre-trained (ALBEF) on image-text pairs shows different patterns, where the blue dots are red dots are somewhat aligned but it does not provide clustered representations. This could be probably due to that the pretext task of ALBEF is to align image and text, but not classification.

\begin{figure}[t]
\centering
    \includegraphics[width=1.0\linewidth]{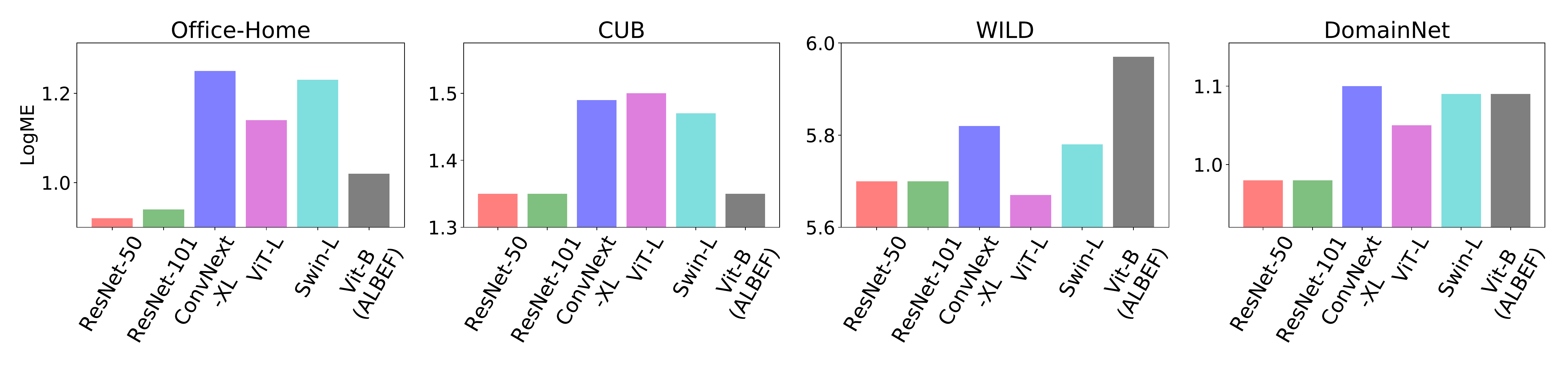}   
    \caption{ Analysis on feature transferability of each pre-trained model. ResNet pre-trained models on ImageNet-1K obtain lower values of LogME than that of state-of-the-art backbones, which implies weak transferability to DA benchmarks}
    \label{fig:logme}
\end{figure}

\noindent\textbf{Analysis on Feature Transferability.} We employ LogME~\cite{you2021logme} to evaluate the transferability of features for downstream tasks. LogME is used to assess pre-trained models, which estimates the maximum value of label evidence from the extracted features of downstream data. A higher value LogME implies better transferability to downstream tasks. We measure LogME of each pre-training for all domains in the benchmarks as shown in Fig.~\ref{fig:logme}. As expected, ResNet backbones pre-trained on ImageNet-1K obtain very low values of LogME compared to the state-of-the-art backbones. We reiterate that the pre-training stage should be modernized according to the recent advances in computer vision.

\section{Conclusions}
Most domain transfer works pay little attention to the importance of the pre-training stage. In this work, we provide an in-depth analysis of the effect of modern pre-training on domain transfer. We summarize some of our key findings:

\begin{enumerate}

\item \textbf{What makes strong pre-training for domain transfer?} In Sec.~\ref{sec:single-source}, we observe that many factors, including network architecture, pre-training dataset, and network size contribute to the improvements in domain transfer tasks. However, there is no single winner across all benchmark datasets. The transferability of pre-training depends on the target benchmark, adaptation method, and network depth. Most importantly, we observe that simply using the SOTA pre-training outperforms all the domain adaptation baselines.

\item \textbf{Do we still need domain adaptation with modern pre-training?} In Sec.~\ref{sec:DA}, while we find that adaptation methods still improve the accuracy with modern pre-training, the relative ranking of domain adaptation methods is not preserved. With modern pre-training, an outdated DA method performs better than more recent DA methods in our experiments.
\end{enumerate}

\noindent \textbf{Limitations.} Due to the availability of pre-trained models, we could not analyze full ablations (\eg, SOTA backbones pre-trained on JFT-300M). In this work, we use a very simple fine-tuning strategy by adding a single FC layer, but there could be other simple better ways to fine-tune pre-trained models for downstream tasks. In addition to image classification tasks, other computer vision tasks including domain adaptive object detection, segmentation, or video domain adaptation should be explored with modern pre-training in future research. We hope future work should use these results as a new baseline. 

\noindent\textbf{Acknowledgments.} This work was supported by DARPA LwLL and NSF Award No. 1535797


\clearpage
%
%
{\small
\bibliographystyle{splncs04}
\bibliography{main}
}

\clearpage

\section{Appendix}
\noindent \textbf{Summary.}
We provide additional details and experimental results in this supplementary material. Future domain adaptation work can use our code\footnote{\url{https://github.com/VisionLearningGroup/Benchmark_Domain_Transfer}} for a new baseline for domain transfer tasks.

\subsection{Additional Training Details}  
\begin{figure}[t]
\centering
    \includegraphics[width=1.0\linewidth]{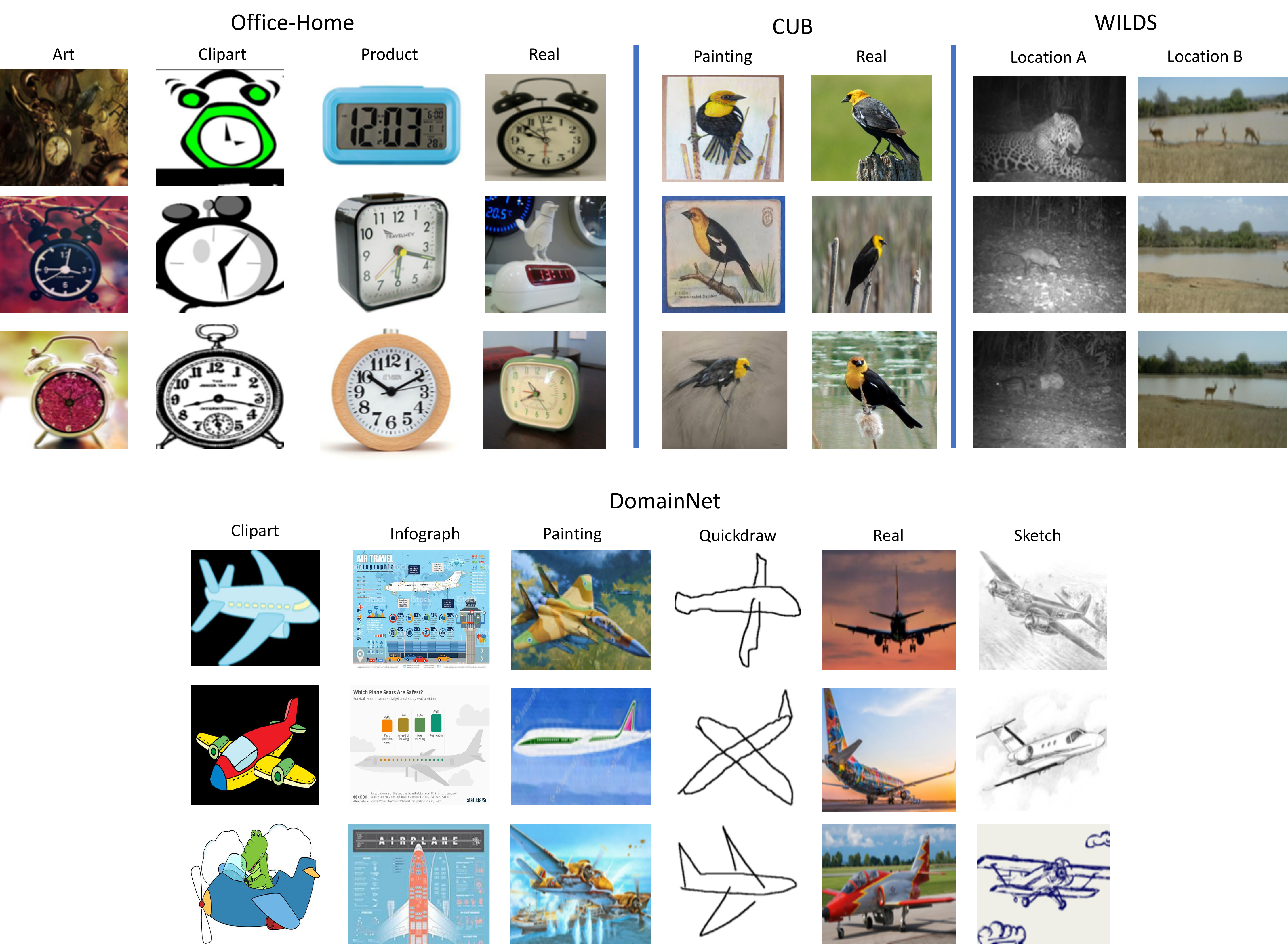}   
    \caption{Examples from different domains in the benchmarks}
    \label{fig:supp_dataset}
\end{figure}

Our implementation is based on the timm\footnote{\url{https://github.com/rwightman/pytorch-image-models}} library and the transfer learning library in~\cite{dalib}. We directly use the implementation of the the transfer learning library in~\cite{dalib}, which supports domain adaptation baselines (DANN, CDAN, MCC, AFN, and MDD). For some pre-trained weights not available in the timm library, we directly use the publicly released pre-trained weights from the authors. In Fig.~\ref{fig:supp_dataset}, we show example images from different domains in each dataset, which show the type of domain shift in each benchmark.

\subsection{Source Accuracy on Single Source Generalization}
\begin{figure}[t]
\centering
    \begin{subfigure}[t]{0.49\textwidth}
        \centering
        \includegraphics[width=1.0\linewidth]{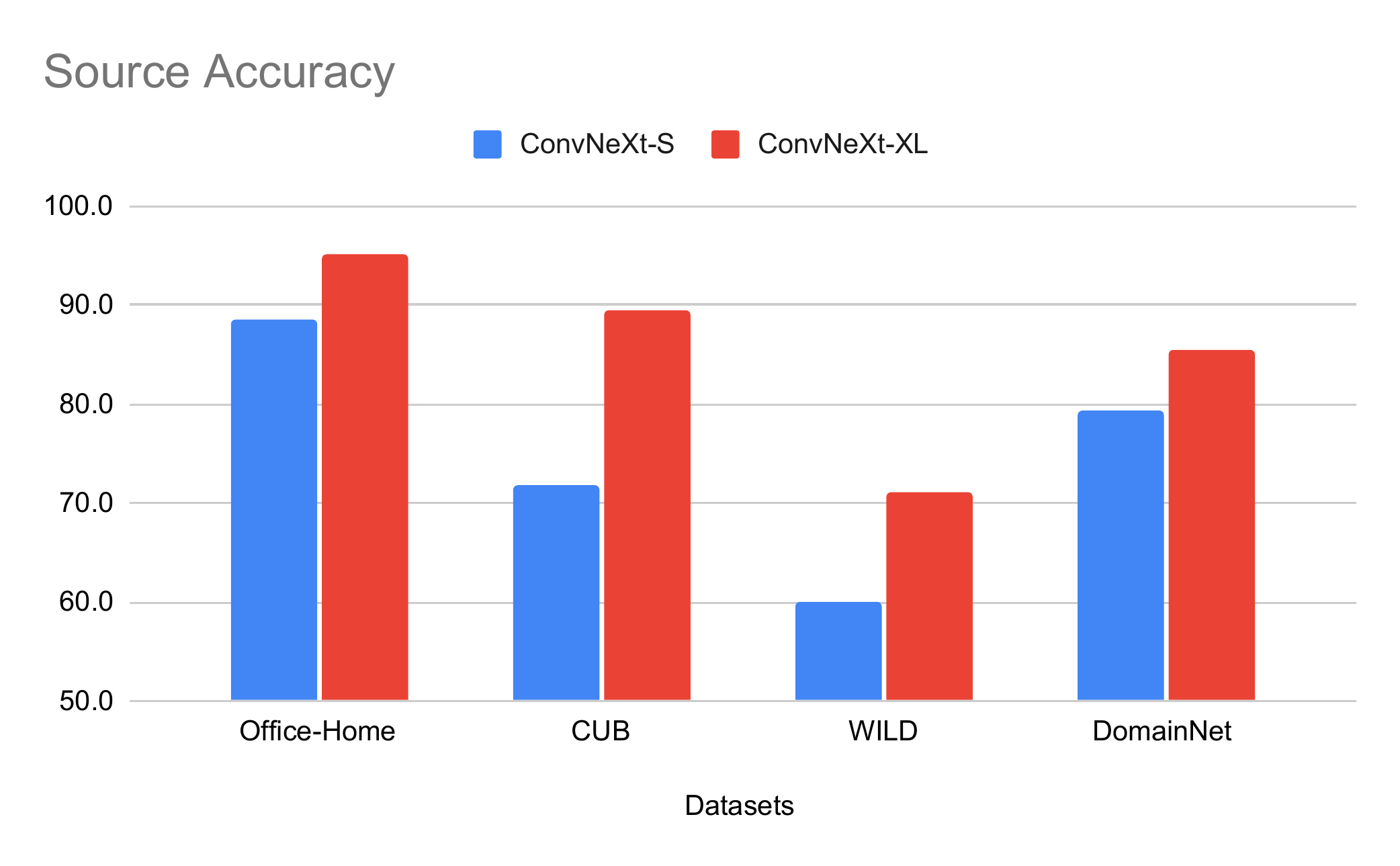}   
        \caption{ConvNeXt}
    \end{subfigure}
    \begin{subfigure}[t]{0.49\textwidth}
        \includegraphics[width=1.0\linewidth]{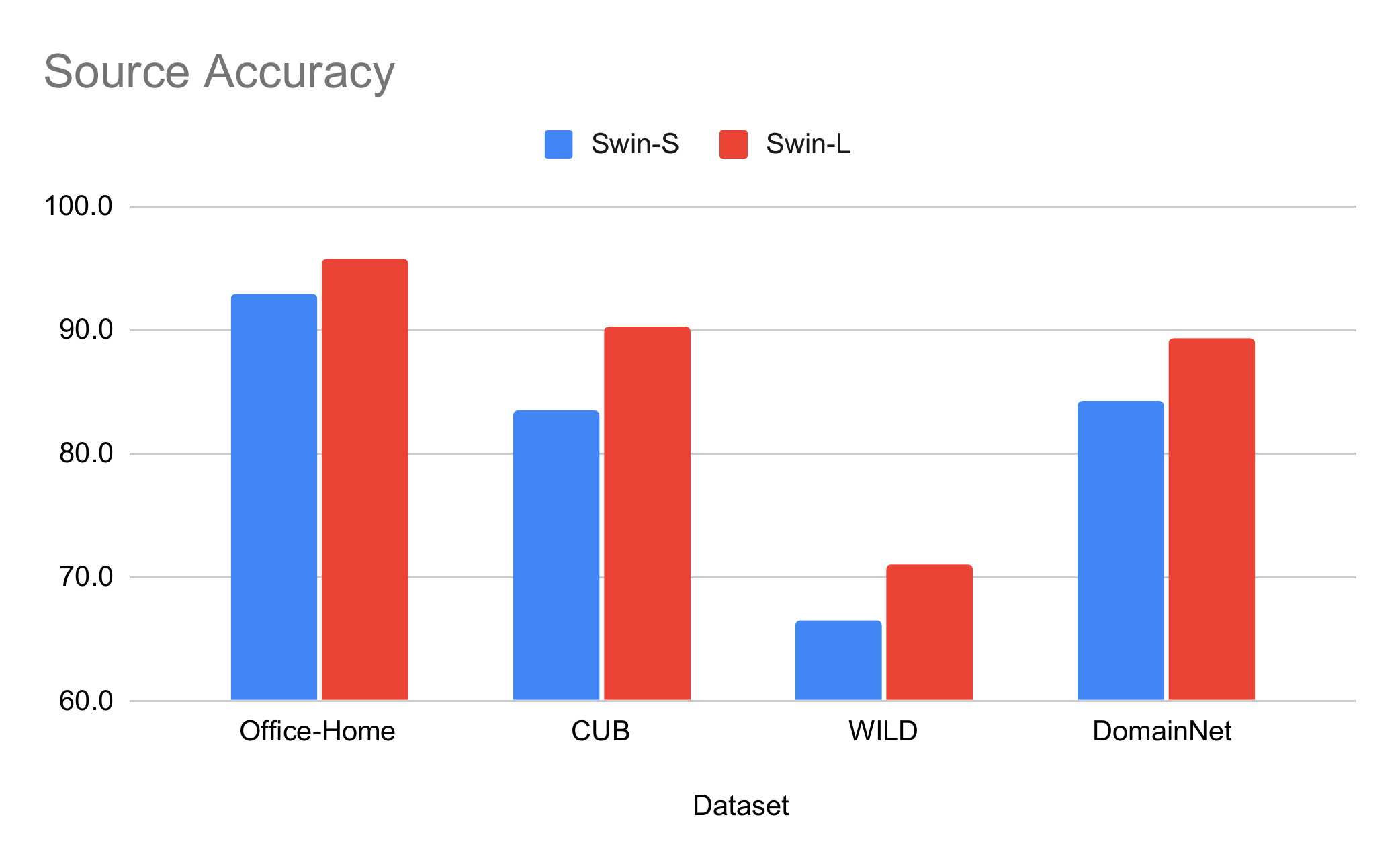}  
        \caption{Swin Transformers}
    \end{subfigure}
    \caption{ (a) Source accuracy comparison on the variants of ConvNeXt (b) Source accuracy comparison on the variants of Swin Transformers
    }
    \label{fig:source_acc}
\end{figure}

We provide a comparison of source accuracy on the source validation set on ConvNext and Swin Transformers in Fig.~\ref{fig:source_acc}. We compare the source accuracy between shallow models, \convnext-S and \swin-S pre-trained on ImageNet-1K, and deep models, \convnext-XL and \swin-L pre-trained on ImageNet-22K. We observe that deep models, \convnext-XL and \swin-L, obtain higher source accuracy on all the benchmarks compared to the shallow models, \convnext-S, \swin-S. According to the theory in in~\cite{ben2010theory}, the  expected target error $\epsilon_{T}(h)$ can be bounded by the  expected source error, the discrepancy between the source and target domains, and the shared error of the ideal joint hypothesis $\lambda$. 
\begin{equation}
\epsilon_{T}(h) \leq  \epsilon_{S}(h)+d_{1}\left(\mathcal{D}_{S}, \mathcal{D}_{T}\right) + \lambda
\end{equation}
Prior domain adaptation methods assume that expected source error $\epsilon_{S}(h)$ is low since we assume many source labels. Therefore, prior domain adaptation methods focus on minimizing the discrepancy between the source and target domains. However, as shown in Fig.~\ref{fig:source_acc}, there is a gap in $\epsilon_{S}(h)$ between pre-trained models. Using modern pre-training can further reduce the upper bound of expected target error with a lower value of $\epsilon_{S}(h)$.

Additionally, we observe that there are noticeable inconsistencies between the source validation accuracy and target accuracy across different optimizers and learning rates in shallow models trained on ImageNet-1K (\eg, ResNet-50, Tiny, Small models of \convnext and \swin) when fine-tuned on CUB and WILD. That means using the higher source validation accuracy for model selection can obtain a very lower target accuracy. We choose the best optimizer and learning rate on target accuracy but use early stopping with source validation accuracy only on these cases for proper comparisons. However, these gaps become smaller in deeper models trained ImageNet-22K and we observe similar trends between source validation and target accuracy. This also suggests that deep models trained on a larger dataset can obtain better domain invariant features than shallow models trained on a smaller dataset.

\begin{table}[t]
\caption{The effect of fine-tuning the text encoder in ALBEF on Office-Home}
\centering
\setlength{\tabcolsep}{4pt}
\resizebox{0.7\textwidth}{!}{
\begin{tabular}{l|c|c|c|c|c}
\toprule[1.0pt]

Backbone & Fine-tune text encoder & Al & Cl & Pr & AVG \\
\hline
ALBEF & \xmark & 81.7 & 72.5 & 87.2 & 80.4 \\
ALBEF & \checkmark & 79.8 & 71.4 & 86.8 & 79.3 \\

\bottomrule[1.0pt]
\end{tabular}}
\label{tab:supp_albef}
\end{table}

\begin{table}[t]
\caption{Additional accuracy comparison on different network architectures and datasets}
\setlength{\tabcolsep}{4pt}
\resizebox{\textwidth}{!}{
\begin{tabular}{l|c|c|c|c|c|c|c|c|c|c|c|c|c}
\toprule[1.0pt]

\multirow{2}{*}{Backbone} & \multirow{2}{*}{Pre-train. Data} & \multirow{2}{*}{Params} & \multicolumn{3}{c|}{Office-Home} & CUB & WILD & \multicolumn{5}{c|}{DomainNet} & \multirow{2}{*}{AVG} \\
\cline{4-13}
 &  &  & Ar & Cl & Pr & Pa & - & Cl & In & Pa & Qu & Sk &  \\
\hline
ResNet-50&	ImageNet-1K&	23 &	66.1&	49.0&	77.2&	42.3&	70.7&	46.6&	17.3&	45.2&	6.5	&	35.3& 50.0 \\
ResNet-101&	ImageNet-1K&	42  &	68.5&	52.4&	79.9&	46.1&	74.0&	49.3&	19.2&	48.6&	8.7&	38.5&	52.9 \\
\deit-B & ImageNet-1K & 85 & 73.4&	54.7&	83.2&	60.6&	74.8&	54.3&	20.7&	51.0&	7.5&	39.5&	56.9\\
\swin-B & ImageNet-1K & 86 & 75.7&	54.7 & 84.8 &	57.1 &	76.6&	56.7&	22.8&	52.6&	8.8	&	41.9& 58.1 \\ 
\convnext-B & ImageNet-1K & 86 & 73.7 &	54.7&	82.4&	45.4&	78.1&	58.2&	24.8&	55.5&	8.5&	45.9&	57.6\\
\hline
ViT-S & ImageNet-22K & 21 & 74.1&	52.7&	83.9&	64.0&	75.7&	53.3&	20.3&	51.2&	7.4&	37.3&	56.9 \\
ViT-B & ImageNet-22K & 85 &	78.4&	57.4&	86.5&	69.9&	76.5&	58.6&	25.0&	57.8&	8.1&	45.3&	61.7 \\
ViT-L & ImageNet-22K & 303  &	84.0&	73.0&	89.9&	76.5&	77.7&	65.5&	27.3&	61.3&	10.2&	52.1&	67.5\\
\swin-B & ImageNet-22K & 86 & 82.6 &	69.1&	90.4	&70.3&	79.7&	63.2&	28.2&	60.0&	10.2&	50.1&	65.9 \\
\swin-L & ImageNet-22K & 194  &83.4&	74.3&	90.9&	73.0&	81.4&	67.2&	30.6&	62.5&	11.2&	54.1&	68.6\\			
\convnext-B & ImageNet-22K & 87 & 81.5	 & 68.0 &	89.9&	65.2&	81.1&	62.7&	26.9&	59.9&	9.4&	52.1&	65.2\\
\convnext-L & ImageNet-22K & 196 &	84.6&	73.2&	90.5&	70.7&	81.1&	66.6&	28.8&	61.6&	10.0&	54.3&	67.9 \\
\convnext-XL & ImageNet-22K & 348 & 85.1&	74.0&	91.4&	71.9&	81.5&	67.7&	29.7&	62.2&	11.4&	55.5&	68.8\\


\bottomrule[1.0pt]
\end{tabular}}
\label{tab:supp_single_source}
\end{table}

\subsection{Additional Results on Single Source Generalization}
In our main paper, we only use the image encoder in ALBEF. We also try to fine-tune the text encoder in ALBEF. Table~\ref{tab:supp_albef} shows that the results of fine-tuning text and image encoder in ALBEF. We observe that fine-tuning the text encoder is not helpful on Office-Home. In Table~\ref{tab:supp_single_source}, we provide an accuracy comparison on different pre-training datasets and network architectures. In Tables~\ref{tab:supp_single_source_office_home} and~\ref{tab:supp_single_source_domainnet}, we provide additional results trained on other source domains on Office-Home and DomainNet.

\begin{table}[t]
\caption{Additional results trained on a different source domain on Office-Home across architectures }
\setlength{\tabcolsep}{4pt}
\resizebox{\textwidth}{!}{
\begin{tabular}{l|c|c|c|c|c|c|c|c|c|c|c|c|c|c|c}
\toprule[1.0pt]

\multirow{2}{*}{Backbone} & \multirow{2}{*}{Pre-train. Data} & \multirow{2}{*}{Params} & \multicolumn{3}{c|}{Source: Ar} & \multicolumn{3}{c|}{Source: Cl} & \multicolumn{3}{c|}{Source: Pr} & \multicolumn{4}{c}{Source: Re} \\
\cline{4-16}
 &  &  & Cl & Pr & Re & Ar & Pr & Re & Ar & Cl & Re & Ar & Cl & Pr & Avg \\
\hline
ResNet-50 & ImageNet-1K & 23 & 46.8 & 64.4 & 71.2 & 52.5 & 62.5 & 63.6 & 49.5 & 42.5 & 72.3 & 66.1 & 49.0 & 77.2 & 58.4 \\
ConvNeXt-T & ImageNet-1K & 27 & 46.2 & 65.0 & 74.0 & 59.3 & 68.3 & 71.8 & 52.1 & 43.3 & 74.8 & 67.4 & 48.7 & 77.9 & 61.6 \\
Swin-T & ImageNet-1K & 27 & 52.3 & 68.8 & 75.5 & 57.9 & 65.9 & 69.3 & 62.1 & 47.6 & 78.8 & 71.3 & 49.4 & 81.1 & 64.2 \\
\hline
ResNet-101 & ImageNet-1K & 42 & 53.3 & 69.5 & 76.1 & 56.5 & 66.2 & 67.0 & 55.2 & 47.1 & 75.1 & 68.5 & 52.4 & 79.9 & 62.9 \\
Swin-S & ImageNet-1K & 48 & 55.0 & 76.8 & 81.9 & 63.3 & 71.6 & 73.7 & 56.5 & 47.6 & 77.9 & 73.8 & 54.5 & 84.2 & 67.1 \\
ConvNeXt-S & ImageNet-1K & 49 & 53.4 & 72.7 & 78.6 & 67.5 & 72.9 & 75.4 & 61.8 & 49.0 & 80.0 & 72.2 & 52.7 & 80.9 & 67.9 \\
\hline
Swin-B & ImageNet-22K & 86 & 70.7 & 86.1 & 88.5 & 80.6 & 84.3 & 86.7 & 77.9 & 66.1 & 88.3 & 82.6 & 69.1 & 90.4 & 81.0 \\
ConvNeXt-B & ImageNet-22K & 87 & 69.7 & 83.0 & 86.9 & 80.5 & 85.9 & 86.5 & 73.8 & 63.2 & 87.6 & 81.5 & 68.0 & 89.9 & 79.7 \\
\hline
Swin-L & ImageNet-22K & 194 & 72.6 & 85.2 & 89.5 & 84.0 & 86.9 & 89.4 & 80.6 & 72.9 & 91.0 & 83.4 & 74.3 & 90.9 & 83.6 \\
ConvNeXt-L & ImageNet-22K & 196 & 71.9 & 86.5 & 89.6 & 83.7 & 85.9 & 88.0 & 80.8 & 67.7 & 89.3 & 84.6 & 73.2 & 90.5 & 82.6 \\
ConvNeXt-XL & ImageNet-22K & 348 & 74.1 & 88.2 & 90.9 & 83.6 & 86.7 & 89.2 & 77.5 & 66.9 & 89.9 & 85.1 & 74.0 & 91.4 & 83.0 \\

\bottomrule[1.0pt]
\end{tabular}}
\label{tab:supp_single_source_office_home}
\end{table}

\begin{table}[t]
\caption{Additional results trained on a different source domain on DomainNet across architectures }
\setlength{\tabcolsep}{4pt}

\begin{subtable}[t!]{1.0\textwidth}
\resizebox{\textwidth}{!}{
\begin{tabular}{l|c|c|c|c|c|c|c|c|c|c|c|c|c|c}
\toprule[1.0pt]
\multirow{2}{*}{Backbone} & \multirow{2}{*}{Pre-train. Data} & \multirow{2}{*}{Params} & \multicolumn{6}{c|}{Source: Cl} & \multicolumn{6}{c}{Source: In} \\
\cline{4-15}
 &  &  & In & Pa & Qu & Re & Sk & AVG & Cl & Pa & Qu & Re & Sk & AVG \\
 \hline
ResNet-50 & ImageNet-1K & 23 & 12.8 & 30.8 & 11.7 & 45.8 & 42.6 & 28.8 & 35.2 & 32.0 & 3.9 & 47.6 & 30.5 & 29.9 \\
Swin-T & ImageNet-1K & 27 & 16.5 & 37.5 & 14.0 & 55.4 & 44.7 & 33.6 & 39.9 & 35.7 & 4.6 & 53.4 & 32.9 & 33.3 \\
ConvNeXt-T & ImageNet-1K & 27 & 16.7 & 39.5 & 13.3 & 57.6 & 46.8 & 34.8 & 40.0 & 37.7 & 4.4 & 54.4 & 34.9 & 34.3 \\
 \hline
ResNet-101 & ImageNet-1K & 42 & 15.7 & 36.2 & 13.2 & 52.6 & 45.3 & 32.6 & 37.9 & 33.4 & 4.9 & 49.6 & 32.1 & 31.6 \\
Swin-S & ImageNet-1K & 48 & 18.9 & 41.5 & 14.6 & 59.8 & 48.3 & 36.6 & 44.6 & 39.4 & 5.9 & 57.2 & 37.8 & 37.0 \\
ConvNeXt-S & ImageNet-1K & 49 & 19.1 & 43.9 & 13.7 & 61.3 & 50.5 & 37.7 & 46.3 & 42.3 & 5.6 & 58.8 & 40.3 & 38.7 \\
 \hline
Swin-B & ImageNet-22K & 86 & 26.2 & 52.8 & 16.7 & 70.5 & 56.8 & 44.6 & 59.2 & 53.0 & 9.9 & 70.9 & 49.7 & 48.5 \\
ConvNeXt-B & ImageNet-22K & 87 & 23.8 & 53.2 & 14.8 & 71.6 & 58.1 & 44.3 & 58.8 & 53.0 & 7.5 & 71.4 & 51.2 & 48.4 \\
 \hline
Swin-L & ImageNet-22K & 194 & 29.5 & 56.5 & 16.9 & 72.9 & 60.1 & 47.2 & 63.5 & 56.1 & 9.7 & 73.6 & 53.3 & 51.2 \\
ConvNeXt-L & ImageNet-22K & 196 & 26.0 & 54.6 & 15.8 & 72.9 & 60.1 & 45.9 & 62.0 & 55.2 & 7.5 & 73.4 & 53.7 & 50.3 \\
ConvNeXt-XL & ImageNet-22K & 348 & 29.2 & 58.3 & 14.6 & 75.5 & 62.2 & 48.0 & 63.7 & 57.2 & 8.1 & 74.7 & 55.1 & 51.8\\
\bottomrule[1.0pt]
\end{tabular}}
\end{subtable}

\begin{subtable}[t!]{1.0\textwidth}
\resizebox{\textwidth}{!}{
\begin{tabular}{l|c|c|c|c|c|c|c|c|c|c|c|c|c|c}
\toprule[1.0pt]
\multirow{2}{*}{Backbone} & \multirow{2}{*}{Pre-train. Data} & \multirow{2}{*}{Params} & \multicolumn{6}{c|}{Source: Pa} & \multicolumn{6}{c}{Source: Qu}  \\
\cline{4-15}
 &  &  & Cl & In & Qu & Re & Sk & AVG & Cl & In & Pa & Qu & Sk & AVG \\
 \hline
ResNet-50 & ImageNet-1K & 23 & 40.7 & 14.5 & 3.0 & 56.2 & 37.3 & 30.4 & 9.4 & 0.7 & 1.2 & 4.1 & 9.1 & 4.9 \\
Swin-T & ImageNet-1K & 27 & 47.5 & 16.7 & 6.8 & 60.4 & 40.6 & 34.4 & 24.9 & 1.9 & 6.2 & 13.3 & 13.8 & 12.0 \\
ConvNeXt-T & ImageNet-1K & 27 & 46.6 & 17.1 & 5.7 & 61.6 & 42.5 & 34.7 & 29.1 & 3.1 & 11.9 & 22.1 & 17.7 & 16.8 \\
 \hline
ResNet-101 & ImageNet-1K & 42 & 44.5 & 16.1 & 4.9 & 57.7 & 39.5 & 32.5 & 9.4 & 0.7 & 1.2 & 4.1 & 9.1 & 4.9 \\
Swin-S & ImageNet-1K & 48 & 50.3 & 19.1 & 6.6 & 62.9 & 43.3 & 36.4 & 28.6 & 2.6 & 8.5 & 17.1 & 16.0 & 14.6 \\
ConvNeXt-S & ImageNet-1K & 49 & 50.7 & 18.8 & 6.0 & 64.3 & 45.2 & 37.0 & 29.1 & 3.3 & 10.3 & 18.9 & 17.8 & 15.9 \\
 \hline
Swin-B & ImageNet-22K & 86 & 60.8 & 25.8 & 7.7 & 72.5 & 51.6 & 43.7 & 32.1 & 3.1 & 9.3 & 18.8 & 19.3 & 16.5 \\
ConvNeXt-B & ImageNet-22K & 87 & 62.3 & 24.9 & 7.7 & 72.9 & 53.7 & 44.3 & 40.3 & 6.3 & 21.2 & 37.3 & 27.1 & 26.4 \\
 \hline
Swin-L & ImageNet-22K & 194 & 64.5 & 28.7 & 9.0 & 75.1 & 53.8 & 46.2 & 40.3 & 6.3 & 20.5 & 35.6 & 26.2 & 25.8 \\
ConvNeXt-L & ImageNet-22K & 196 & 64.5 & 28.6 & 8.9 & 75.0 & 53.4 & 46.1 & 42.0 & 7.8 & 24.0 & 39.5 & 28.0 & 28.3 \\
ConvNeXt-XL & ImageNet-22K & 348 & 64.1 & 27.5 & 7.9 & 75.7 & 56.3 & 46.3 & 43.2 & 8.2 & 27.1 & 43.0 & 29.7 & 30.3\\
\bottomrule[1.0pt]
\end{tabular}}
\end{subtable}

\begin{subtable}[t!]{1.0\textwidth}
\resizebox{\textwidth}{!}{
\begin{tabular}{l|c|c|c|c|c|c|c|c|c|c|c|c|c|c}
\toprule[1.0pt]
\multirow{2}{*}{Backbone} & \multirow{2}{*}{Pre-train. Data} & \multirow{2}{*}{Params} & \multicolumn{6}{c|}{Source: Re} & \multicolumn{6}{c}{Source: Sk}  \\
\cline{4-15}
 &  &  & Cl & In & Pa & Qu & Sk & AVG & Cl & In & Pa & Qu & Re & AVG \\
 \hline
ResNet-50 & ImageNet-1K & 23 & 46.6 & 17.3 & 45.2 & 6.5 & 35.3 & 30.2 & 52.4 & 13.3 & 36.3 & 12.3 & 47.5 & 32.3 \\
Swin-T & ImageNet-1K & 27 & 50.8 & 20.6 & 50.8 & 7.8 & 41.2 & 34.2 & 56.8 & 15.0 & 39.3 & 14.5 & 50.7 & 35.3 \\
ConvNeXt-T & ImageNet-1K & 27 & 49.3 & 19.8 & 37.3 & 7.3 & 37.3 & 30.2 & 56.6 & 15.1 & 41.5 & 13.7 & 53.5 & 36.1 \\
 \hline
ResNet-101 & ImageNet-1K & 42 & 49.3 & 19.2 & 48.6 & 8.7 & 38.5 & 21.8 & 53.8 & 12.9 & 35.5 & 13.7 & 43.2 & 31.8 \\
Swin-S & ImageNet-1K & 48 & 55.9 & 22.5 & 51.8 & 8.6 & 41.4 & 36.0 & 60.2 & 16.9 & 42.5 & 15.5 & 54.4 & 37.9 \\
ConvNeXt-S & ImageNet-1K & 49 & 54.9 & 22.2 & 52.8 & 8.1 & 43.0 & 36.2 & 60.7 & 17.6 & 44.6 & 14.1 & 57.6 & 38.9 \\
 \hline
Swin-B & ImageNet-22K & 86 & 63.2 & 28.2 & 60.0 & 10.2 & 50.1 & 42.3 & 68.4 & 24.1 & 53.9 & 17.6 & 68.4 & 46.5 \\
ConvNeXt-B & ImageNet-22K & 87 & 62.7 & 26.9 & 59.9 & 9.4 & 52.1 & 42.2 & 68.9 & 22.3 & 54.4 & 15.0 & 68.4 & 45.8 \\
 \hline
Swin-L & ImageNet-22K & 194 & 67.2 & 30.6 & 62.5 & 11.2 & 54.1 & 45.1 & 68.4 & 26.6 & 54.9 & 17.5 & 69.8 & 47.5 \\
ConvNeXt-L & ImageNet-22K & 196 & 66.6 & 28.8 & 61.6 & 10.0 & 54.3 & 44.3 & 69.9 & 24.2 & 56.4 & 16.6 & 69.8 & 47.4 \\
ConvNeXt-XL & ImageNet-22K & 348 & 67.7 & 29.7 & 62.2 & 11.4 & 55.5 & 45.3 & 70.4 & 26.0 & 58.2 & 16.4 & 73.5 & 48.9 \\
\bottomrule[1.0pt]
\end{tabular}}
\end{subtable}

\label{tab:supp_single_source_domainnet}
\end{table}

\end{document}